\documentclass[a4paper,num-refs]{oup-contemporary-arxiv}

\usepackage{graphicx}
\usepackage{siunitx}
\usepackage{multirow}

\title{ChronoRoot 2.0: An Open AI-Powered Platform for 2D Temporal Plant Phenotyping}

\author[1,2,3,4,5, \authfn{1}]{Nicolás Gaggion}
\author[1, 2]{Noelia A. Boccardo}
\author[6]{Rodrigo Bonazzola}
\author[1,2]{María Florencia Legascue}
\author[2]{María Florencia Mammarella}
\author[1]{Florencia Sol Rodriguez}
\author[1]{Federico Emanuel Aballay}
\author[1]{Florencia Belén Catulo}
\author[4,5]{Andana Barrios}
\author[1]{Luciano J. Santoro}
\author[2]{Franco Accavallo}
\author[1,2]{Santiago Nahuel Villarreal}
\author[4,5]{Leonardo I. Pereyra-Bistrain}
\author[4, 5]{Moussa Benhamed}
\author[4, 5]{Martin Crespi}
\author[1]{Martiniano María Ricardi}
\author[1]{Ezequiel Petrillo}
\author[4, 5]{Thomas Blein}
\author[1]{Federico Ariel}
\author[3]{Enzo Ferrante}

\affil[1]{Instituto de Fisiología, Biología Molecular y Neurociencias (IFIBYNE), CONICET-Universidad de Buenos Aires, Argentina}
\affil[2]{APOLO Biotech, Argentina}
\affil[3]{Instituto de Ciencias de la Computación, CONICET-Universidad de Buenos Aires, Argentina}
\affil[4]{Université Paris-Saclay, CNRS, INRAE, Université Evry, Institute of Plant Sciences Paris-Saclay (IPS2), 91190 Gif-sur-Yvette, France}
\affil[5]{Université Paris Cité, CNRS, INRAE, Institute of Plant Sciences Paris-Saclay (IPS2), 91190 Gif-sur-Yvette, France}
\affil[6]{Instituto de Investigación en Señales, Sistemas e Inteligencia Computacional sinc(i), CONICET-Universidad Nacional del Litoral, Argentina}

\authnote{\authfn{1}Address correspondence to: ngaggion@dc.uba.ar}

\papercat{Technical Note}

\runningauthor{Gaggion et al.}

\jvolume{00}
\jnumber{0}
\jyear{2026}

\begin{document}

\begin{frontmatter}

\maketitle

\begin{abstract}

\noindent \textbf{Background:} Plant developmental plasticity, particularly in root system architecture, is fundamental to understanding adaptability and agricultural sustainability. Existing automated phenotyping solutions face limitations including binary segmentation approaches, restricted structural analysis capabilities, and text-based interfaces that limit accessibility, with most focusing solely on root structures while overlooking valuable information from simultaneous analysis of multiple plant organs.

\noindent \textbf{Findings:} ChronoRoot 2.0 builds upon established low-cost hardware while significantly enhancing software capabilities and usability. The system employs nnUNet architecture for multi-class segmentation, demonstrating significant accuracy improvements while simultaneously tracking six distinct plant structures encompassing root, shoot, and seed components: main root, lateral roots, seed, hypocotyl, leaves, and petiole. This architecture enables easy retraining and incorporation of additional training data without requiring machine learning expertise. The platform introduces dual specialized graphical interfaces: a Standard Interface for detailed architectural analysis with novel gravitropic response parameters, and a Screening Interface enabling high-throughput analysis of multiple plants through automated tracking. Functional Principal Component Analysis integration enables discovery of novel phenotypic parameters through temporal pattern comparison. \textcolor{black}{We demonstrate multi-species analysis, with \textit{Arabidopsis thaliana} and \textit{Solanum lycopersicum}, both morphologically distinct plant species.} Three use cases in \textit{Arabidopsis thaliana} \textcolor{black}{and validation with tomato seedlings} demonstrate enhanced capabilities: circadian growth pattern characterization, gravitropic response analysis in transgenic plants, and high-throughput etiolation screening across multiple genotypes.

\noindent \textbf{Conclusions:} ChronoRoot 2.0 maintains the low-cost, modular hardware advantages of its predecessor while dramatically improving accessibility through intuitive graphical interfaces and expanded analytical capabilities. The open-source platform makes sophisticated temporal plant phenotyping more accessible to researchers without computational expertise.

\noindent \textbf{Software availability:} https://chronoroot.github.io

\end{abstract}


\end{frontmatter}

\section{Introduction}

Plants, as sessile organisms, must develop sophisticated adaptive strategies to cope with their immediate environment throughout their lifecycle. This fundamental biological constraint has driven the evolution of remarkable developmental plasticity, enabling plants to complete their life cycles under varying and often suboptimal growth conditions \cite{John_Wiley_Sons_Ltd2001-ze}. The root system, being the primary interface between plant and soil, exhibits particularly notable phenotypic plasticity in response to environmental variables \cite{Tracy2020-fw}. Understanding the dynamics of root system development and its plastic responses has become increasingly critical in the context of climate change and the growing need for sustainable agriculture.

Under controlled conditions, root development is typically observed through images of plants growing vertically on semisolid agarized medium. Root system architecture (RSA) is then characterized through various parameters such as main root length and lateral root density \cite{Ingram2010-vy}. Several semi-automatic tools assist in root phenotyping at specific time points, yet comprehensive temporal analysis remains technologically challenging \cite{Narisetti2019-sd, Yasrab2019-py}. In addition,  most existing tools focus exclusively on root structures, overlooking valuable information that could be gained from analyzing other plant organs simultaneously, from an integrative perspective.

The original ChronoRoot \cite{gaggion2021chronoroot} system introduced automated temporal phenotyping through a low-cost approach combining off-the-shelf electronics, 3D printed hardware components, and deep learning models. This system demonstrated the potential for automated analysis in plant root developmental studies through high-throughput temporal phenotyping of \textit{Arabidopsis thaliana} RSA. However, its practical application revealed several limitations that restricted its broader adoption in the plant science community.

The binary segmentation approach of the original Chronoroot, while effective for basic root architecture analysis, proved inadequate for capturing the full complexity of plant development, particularly during early growth stages. The requirement for manual seed positioning created a bottleneck in high-throughput analysis, while the text-based interface presented a barrier to adoption by researchers without computational expertise. Additionally, the system's focus on root structures alone meant that valuable information about other plant organs and their developmental relationships was not captured. These limitations highlighted the need for a more comprehensive and accessible solution that could capture the full complexity of plant development while maintaining analytical rigor. 

\subsection{ChronoRoot 2.0: An Integrated Solution for Plant Phenotyping}
\begin{figure*}[t!]
\centering
\includegraphics[width=1.0\linewidth]{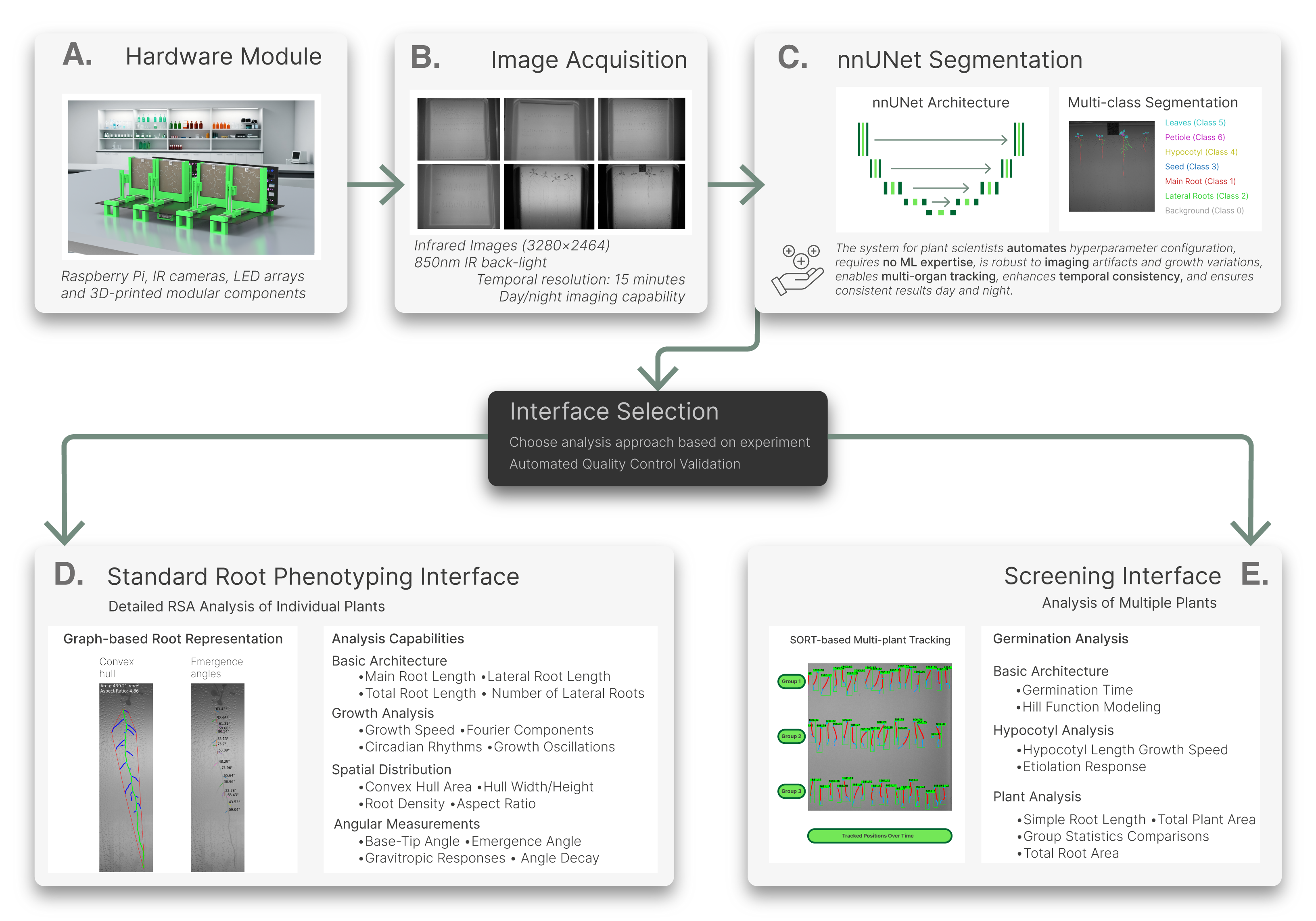}
\caption{\textbf{ChronoRoot 2.0: An integrated platform for temporal plant phenotyping.} (A) The hardware module combines affordable components for automated imaging in controlled environments. (B) Infrared images are captured continuously, enabling consistent monitoring during both day and night cycles. (C) A multi-class segmentation model based on nnUNet automatically identifies and tracks six plant structures: main root, lateral roots, seed, hypocotyl, leaves, and petiole. The system routes data through an interface selection step, offering two specialized analysis workflows: (D) the Standard Interface for detailed architectural analysis of individual plants, and (E) the Screening Interface for high-throughput experiments involving multiple individuals.}
\label{fig:fig1}
\end{figure*}

Building upon this foundation, we present ChronoRoot 2.0, which significantly expands the capabilities and accessibility of the platform through several key innovations, as illustrated in the comprehensive pipeline shown in Figure \ref{fig:fig1}. The complete workflow begins with the Hardware Module (Fig. \ref{fig:fig1}-A) that combines Raspberry Pi-controlled cameras and infrared LED backlighting. This setup enables consistent Image Acquisition (Fig. \ref{fig:fig1}-B) irrespective of day/night cycles, with a temporal resolution of 15 minutes for extended monitoring of plant development.

At the core of these improvements is an upgraded nnUNet Segmentation module (Fig. \ref{fig:fig1}-C) based on the nnUNet architecture \cite{isensee2021nnu}, which performs simultaneous multi-class segmentation of six distinct plant structures: main root (class 1), lateral roots (class 2), seed (class 3), hypocotyl (class 4), leaves (class 5) and petiole (class 6). This advancement enables comprehensive tracking of plant development from seed to mature seedling, capturing the intricate relationships between different organs during growth. The choice of nnUNet as the core architecture was motivated by its proven success in medical image segmentation and its ability to self-configure hyperparameters, making it particularly suitable for biologists without extensive machine learning expertise. While the segmentation model is trained to work with \textcolor{black}{both} \textit{Arabidopsis thaliana} \textcolor{black}{and \textit{Solanum lycopersicum} (tomato)}, it \textcolor{black}{can be} easily \textcolor{black}{adapted} for other species\textcolor{black}{, leveraging the self-configuring capabilities of the nnUNet architecture}. This architectural choice, combined with newly developed graphical user interfaces, significantly lowers the barrier to entry for researchers seeking to implement plant phenotyping in their work.

After segmentation, researchers can select between two distinct but complementary interfaces based on their experimental needs. The Standard Root Phenotyping Interface (Fig. \ref{fig:fig1}-D) maintains continuity with the original ChronoRoot design, focusing on detailed RSA analysis of individual plants. This interface provides researchers with tools for precise measurement and analysis of root development patterns through a graph-based representation approach. It maintains the core strengths of the original ChronoRoot system while adding comprehensive visualization capabilities and an intuitive graphical user interface for analyzing basic architecture, growth dynamics, spatial distribution, and newly determined angular measurements.

The new Screening Interface (Fig. \ref{fig:fig1}-E) extends the system's capabilities by enabling automated analysis of multiple plants simultaneously. It incorporates the Simple Online Realtime Tracking (SORT) algorithm\cite{bewley2016sort} for robust plant identification across experimental groups, along with manual calibration tools for standardized measurements. This interface specializes in early development analysis through three dedicated modules: germination analysis, hypocotyl analysis, and plant analysis, enabling researchers to efficiently process and compare multiple experimental conditions.

Both interfaces implement comprehensive quality control mechanisms through real-time feedback via interactive visualization tools. The system introduces several novel analytical capabilities, including automated seed detection and tracking that eliminates the need for manual positioning. Remarkably, we have incorporated here Functional Principal Component Analysis (FPCA) for time series comparison across different groups of plants (e.g. genotypes, treatments, or combinations), enabling the discovery of new data-driven phenotypic parameters that may not be apparent through traditional analysis methods.

Through this dual-interface approach, ChronoRoot 2.0 addresses diverse needs of the plant science community. The Standard Interface provides the precise control and detailed analysis capabilities needed for in-depth root architecture studies of individual plants, while the Screening Interface enables efficient processing of multiple plants and experimental groups, making high-throughput phenotyping accessible to a broader research community.

The remainder of this paper details the technical implementations and validations of each component of Chronoroot 2.0. We begin by describing the enhanced segmentation capabilities and their validation, followed by detailed explanations of the specialized analyses enabled by each interface. We then present the statistical frameworks implemented for both detailed RSA studies and multi-plant screening experiments. Finally, we discuss the system's graphical user interfaces and their role in making advanced phenotyping accessible to the broader plant science community.


\section{Materials and Methods}

\subsection{System Architecture Overview}

ChronoRoot 2.0 is built as a modular software system that integrates hardware control, image processing, and analysis capabilities within a unified framework. The system architecture comprises three main components: a hardware control module for image acquisition using fixed-focus infrared cameras, a deep learning-based segmentation core for multi-class plant structure identification, and two specialized graphical user interfaces designed for different experimental scenarios.

The components operate through a standardized data pipeline: the hardware module captures infrared images at defined intervals, which are then processed by the segmentation core to identify distinct plant structures, before being analyzed through either of the specialized interfaces. The Standard Root Phenotyping Interface enables detailed architectural analysis through graph-based representation, while the Screening Interface facilitates high-throughput analysis of multiple plants through automated plant tracking. This integrated architecture ensures consistent data processing while enabling flexible deployment for various experimental needs.

\subsection{Hardware Implementation and Image Acquisition}

The ChronoRoot device \cite{gaggion2021chronoroot} is an affordable and modular imaging system based on 3D-printed and laser cut pieces combined with off-the-shelf electronics. Each module consists of a Raspberry Pi (v3)-embedded computer controlling four fixed-zoom and fixed-focus cameras (RaspiCam v2), and an array of infrared (IR) LED back-light. In between each camera and the corresponding IR array, there is a vertical 12 x 12 cm plate for seedling growth, allowing automatic image acquisition repeatedly along the experiment without any modification or movement of the imaging setup.

The four-plate module is compact (62 x 36 x 20 cm) and can be placed in any standard plant growth chamber. The different parts of the imaging setup (back-light, plate support and camera) can be positioned along a horizontal double-rail to control the field of view of the camera and accurate lighting. In addition, the camera can be moved vertically. ChronoRoot allows image acquisition at a high temporal resolution (a set of pictures every minute). The use of an IR back-light (850 nm) and optional long pass IR filters ($>$ 830 nm) allow acquiring images of the same quality independently from the light conditions required for the experiment, during day and night.

\subsection{Deep Learning Segmentation Framework}

\textcolor{black}{ChronoRoot 2.0's machine learning capabilities rests on a comprehensive dataset of plant developmental sequences that represents a significant expansion from the original ChronoRoot. The dataset comprises 911 manually annotated images of \textit{Arabidopsis thaliana} and 480 images of tomato, capturing multiple plants across various developmental stages. Expert biologists performed detailed annotation using ITK-SNAP \cite{yushkevich2016itk}, chosen for its precise annotation tools and user-friendly interface.}

The annotation process captured seven distinct structural classes:
\begin{itemize}
\item Class 0: Background (non-plant regions)
\item Class 1: Main root (primary root axis)
\item Class 2: Lateral roots (all secondary roots)
\item Class 3: Seed (pre- and post-germination structures)
\item Class 4: Hypocotyl (stem region between root-shoot junction and cotyledons)
\item Class 5: Leaves (including both cotyledons and true leaves)
\item Class 6: Petiole (stalk that attaches the leaf to the stem)
\end{itemize}

\textcolor{black}{In the tomato dataset, leaves and petiole were annotated as a single combined aerial part class, reflecting species-specific morphological differences.}

Image acquisition utilized the ChronoRoot hardware system's \textcolor{black}{Raspberry Pi Noir V2 camera}, producing infrared images at 3280 x 2464 resolution \textcolor{black}{(approximately $\approx$ 0.04 mm per pixel)}. \textcolor{black}{Images are processed at single channel full native resolution, preserving all structural detail throughout both training and inference.} 
To ensure robust model generalization, the dataset incorporates specimens from various experimental conditions and genetic backgrounds.

The segmentation framework implements the nnUNet architecture \cite{isensee2021nnu}, selected for its self-configuring capabilities and proven performance in biomedical image analysis. The network dynamically adjusts its depth, width, kernel sizes, stride patterns, normalization schemes, and learning rate schedules based on the training dataset's properties, eliminating the need for extensive hyperparameter tuning. This adaptability is particularly valuable in plant phenotyping contexts, where morphological diversity and varying experimental conditions create unique imaging challenges. \textcolor{black}{Our implementation uses the official nnUNet v2 framework, facilitating straightforward adaptation to new plant species through standard retraining procedures. The framework provides two architectural variants: the standard convolutional nnUNet and the recently introduced nnUNet with residual encoder connections \cite{nnunetrevisited}, which improves feature learning through skip connections in the encoder pathway. Additionally, the framework supports optional test-time augmentation, where predictions from multiple augmented versions of each image are averaged to improve segmentation robustness, particularly for boundary detection.}

\subsection{Specialized Processing Pipelines}

ChronoRoot 2.0 implements a multi-stage processing pipeline that begins with temporal consistency enhancement of the segmentation outputs. All segmentations produced by the nnUNet undergo a weighted trailing average approach, with special consideration for the multi-class nature of the predictions. This temporal integration strategy significantly improves tracking robustness by incorporating historical structural information alongside new observations. This temporal averaging is selectively applied only to the main root (class 1) and lateral roots (class 2) channels, as these structures require particular stability for accurate tracking. The accumulation is expressed as $a^t = s^t + \alpha a^{t-1}$, where $s^t$ is the current segmentation at time $t$, $a^{t-1}$ is the accumulated mask up to the previous time step, and $\alpha$ is a weight factor determined by the temporal resolution of the sequence. This approach effectively addresses common imaging challenges in plant phenotyping, such as water droplets, condensation artifacts, or temporary occlusions, providing stable root structure detection throughout developmental timeframes.

Following this temporal processing, the \textbf{Standard Root Phenotyping Interface} implements a Region of Interest (ROI)-based analysis approach similar to the one available in the original ChronoRoot system. After initial segmentation of the full image, this interface requires user interaction to define individual ROIs for each plant to be analyzed. This manual ROI selection is crucial for ensuring accurate, independent processing without interference from neighboring specimens. Within each ROI, the system first performs binary mask refinement through morphological operations and connected component analysis. The subsequent skeletonization and graph construction processes operate solely on the refined binary mask within the current ROI, ensuring that the resulting graph structure represents only the selected plant's root system. This approach enables precise measurement of root system architecture parameters and growth patterns while maintaining the ability to analyze multiple plants from the same image sequence over time through sequential processing.

The \textbf{Screening Interface} extends the system's capabilities to high-throughput scenarios through automated multi-plant tracking based on the SORT (Simple Online Realtime Tracking) algorithm \cite{bewley2016sort}. The tracking system begins with robust seed detection through contour analysis of segmentation masks and maintains plant identities across frames through a sophisticated combination of Kalman filtering \cite{kalman} and the Hungarian \cite{hungarian} algorithm. Kalman filtering enables prediction of plant positions in subsequent frames based on their movement patterns, while the Hungarian algorithm optimizes the association between predicted and detected positions, ensuring reliable tracking even in crowded scenes. The interface supports definition of experimental groups for comparative studies and implements comprehensive quality control mechanisms during data postprocessing, including automatic removal of plants that cover or touch each other or exhibit abnormal movement patterns. This automated approach enables simultaneous but simpler analysis of multiple plants while maintaining measurement accuracy, significantly increasing experimental throughput without compromising data quality.

\subsection{Analysis Frameworks}

The analysis capabilities of ChronoRoot 2.0 comprise three main components: the Standard Root System Architecture Analysis that maintains continuity with the original ChronoRoot system while adding enhanced features, the High-Throughput Screening Analysis that enables efficient processing of multiple plants simultaneously, and a new module implementing Functional Principal Component Analysis (FPCA) that provides sophisticated temporal pattern analysis. In what follows we provide a more detailed description of each module.

\subsubsection{Standard Root System Architecture Analysis}

The Standard Root Phenotyping Interface provides detailed architectural analysis of individual plant root systems through time. This analysis pipeline builds upon the core capabilities of the original ChronoRoot system while introducing new measurements and enhanced processing methods.

The analysis begins with user definition of ROIs, allowing precise selection of individual plants from multi-plant images. Within these ROIs, the system processes segmentation masks through morphological operations and thinning algorithms to obtain skeletal representations of the root system. These skeletons are then analyzed to identify key nodes, which serve as the basis for constructing a graph representation using a depth-first search algorithm \cite{Cormen2001-jv}.

The graph-based approach, combined with temporal tracking of nodes across frames, enables automatic both node and edge classification and measurement of key architectural features. The system distinguishes between main root and lateral root segments through analysis of the graph structure, with special consideration for complex topologies such as loops where lateral roots reconnect with the main root axis. All measurements provided by ChronoRoot 2.0 are summarized in Table \ref{tab:metrics}, organized into five categories: basic architecture, growth analysis, spatial distribution, angular measurements, and high-throughput analysis. The rightmost column indicates which use case (numbered 1-3) demonstrates the practical application of each metric, with detailed results presented in the Results section.

Basic architectural parameters capture the fundamental dimensions of the root system through main root length, total lateral root length, and their relationships. Growth dynamics are analyzed through temporal derivatives of these measurements, with special attention to circadian patterns revealed through Fourier analysis of filtered growth speeds. Spatial distribution metrics, computed daily, use convex hull analysis to characterize the overall root system shape and space utilization.

Building upon these established measurements, ChronoRoot 2.0 introduces novel angular parameters that provide detailed insight into lateral root development patterns. These measurements are particularly relevant for quantifying gravitropic responses and directional growth dynamics that are central to plant developmental studies. The measurements leverage both the graph structure, which provides precise identification of lateral root base and tip positions, and the labeled skeleton representation, which enables tracking of root paths for emergence angle calculations.

Two complementary angles are calculated with respect to the vertical axis, where 0 degrees represents perfectly vertical growth. The base-tip angle ($\theta_{bt}$) measures the overall orientation using three reference points: the root base coordinates ($x_b, y_b$), the root tip coordinates ($x_t, y_t$), and the vertical projection of the tip ($x_b, y_t$):

\begin{equation}
\theta_{bt} = \arccos\left(\frac{y_t - y_b}{\sqrt{(x_t - x_b)^2 + (y_t - y_b)^2}}\right) \cdot \frac{180}{\pi}
\end{equation}

This measurement captures the terminal orientation of the root after all developmental adjustments have occurred. In contrast, the emergence angle ($\theta_e$) quantifies the initial growth trajectory by measuring the angle at a fixed distance $d$ (default 2 mm) from the base:

\begin{equation}
\theta_e = \arccos\left(\frac{y_d - y_b}{\sqrt{(x_d - x_b)^2 + (y_d - y_b)^2}}\right) \cdot \frac{180}{\pi}
\end{equation}

\begin{figure}
    \centering
    \includegraphics[width=\linewidth]{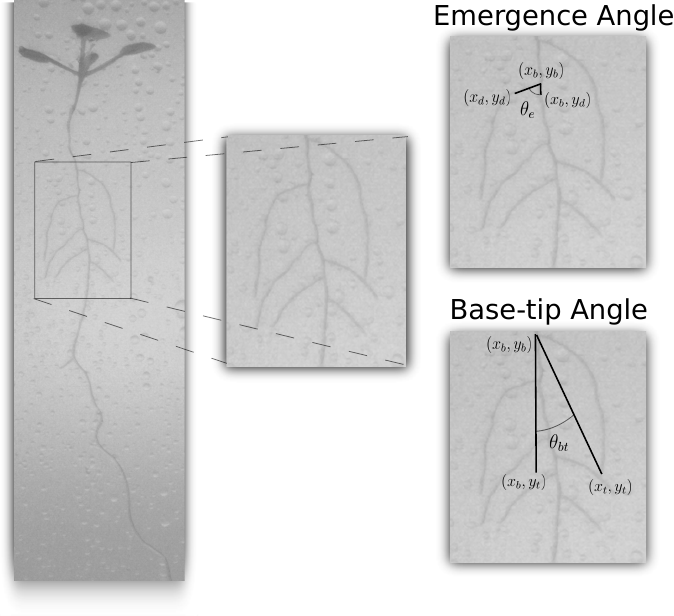}
    \caption{\textbf{Angular measurements in root system architecture.} Illustration of base-tip angle ($\theta_{bt}$) and emergence angle ($\theta_{e}$) calculations \textcolor{black}{on a \textit{Arabidopsis thaliana} plant} showing how these complementary metrics quantify different aspects of lateral root orientation.}
    \label{fig:angles}
\end{figure}

For these novel angular parameters (illustrated in Figure \ref{fig:angles}), the system maintains temporal consistency through careful tracking of individual root components between frames. This temporal integration is particularly important for lateral roots, whose identities must be preserved across timepoints to enable reliable measurement of architectural changes throughout development.

To facilitate data exchange and integration with the broader plant phenotyping community, all measurements are exported in the Root System Markup Language (RSML) format \cite{RSML}. This standardized XML-based format stores the complete hierarchical structure of main and lateral roots along with their spatial coordinates and derived metrics, enabling interoperability with other root phenotyping tools. 

\textcolor{black}{To prepare the data for final analysis and visualization, all measurements pass through an automated post-processing script that ensures biological and structural consistency across the time series. To account for skeletonization noise and avoid the inclusion of false-positive "spurs" on the root axes, we implemented a structural pruning threshold where a minimum length of 5 pixels ($\approx$ 0.2 mm) is required for a skeleton branch to be preserved prior to graph construction. Additionally, a measurement post-processing script filters out "false starts" and transient misdetections through a temporal verification window: any detected structure is only validated if it persists beyond a 6-hour threshold. Finally, the pipeline enforces a monotonic growth constraint for length measurements, preventing impossible decreases in plant size over time. While these cleaned results are used for the study figures, the raw segmentations, graphs, and original RSML data remain available for researchers who wish to perform their own specific analyses.}

\begin{table*}[t!]
\centering
\caption{\textbf{Overview of metrics provided by ChronoRoot 2.0.} Measurements organized by category: Basic Architecture, Growth Analysis, Spatial Distribution, Angular Measurements, and High-throughput Analysis. The Use Case column indicates which demonstration study (\#1: circadian analysis, \#2: gravitropic response, \#3: etiolation screening) employs each metric.}
\resizebox{0.99\linewidth}{!}{
\begin{tabular}{llllc}
\toprule
\textbf{Category} & \textbf{Metric} & \textbf{Units} & \textbf{Measurement Method} & \textbf{Use Case} \\
\midrule
\multirow{7}{*}{Basic Architecture} 
& Main Root (MR) Length & mm & Path length along graph skeleton & 1 \\ 
& Lateral Root (LR) Length & mm & Sum of all LR path lengths & 1 \\ 
& Total Root (TR) Length & mm & MR Length + LR Length & 1 \\ 
& Number of Lateral Roots & count & Unique edges emerging from main root path & 1 \\ 
& Discrete LR Density & LRs/cm & 10 * Number of LRs / MR Length & 1 \\ 
& Main Over Total Root & ratio & MR Length / TR Length & 1 \\ 
\midrule
\multirow{3}{*}{Growth Analysis\textsuperscript{*}} 
& Growth Speed & mm/h & Temporal derivative of length measurements & 1 \\
& Detrended Growth Speeds & mm/h & Raw speeds minus median-filtered signal & 1 \\
& Fourier Components & Hz & Fast Fourier Transform of detrended time series & 1 \\
\midrule
\multirow{5}{*}{Spatial Distribution} 
& Convex Hull Area & mm² & OpenCV convex hull function & 2 \\ 
& Convex Hull Width & mm & Maximum horizontal extent & 2 \\ 
& Convex Hull Height & mm & Maximum vertical extent & 2 \\ 
& Root Density & mm/mm² & Root Length / Convex Hull Area & 2 \\ 
& Aspect Ratio & ratio & Height / Width & 2 \\ 
\midrule
\multirow{2}{*}{Angular Measurements} 
& Base-Tip Angle & degrees & Angle between vertical and root tip & 2 \\ 
& Emergence Angle & degrees & Angle at 2mm from base & 2 \\ 
\midrule
\multirow{9}{*}{Multiple Plant Analysis} 
& Germination Time & hours & Time to radicle emergence & 3 \\
& T50 & hours & Time to 50\% germination & 3 \\
& TMGR & hours & Time of maximum germination rate & 3 \\
& Final Germination & \% & Percentage of germinated seeds & 3 \\
& Seed Size & mm² & Area at experiment start & 3 \\
& Hypocotyl Length & mm & Path length of hypocotyl skeleton & 3 \\
& Hypocotyl Growth Speed & mm/h & Length difference between timepoints & 3 \\
& Total Plant Area & mm² & Sum of all segmented regions & 3 \\
& Simple Root Length & mm & Path length of primary root skeleton & 3 \\
\bottomrule
\multicolumn{5}{l}{\textsuperscript{*}Applicable to any temporal measurement} \\
\end{tabular}}
\label{tab:metrics}
\end{table*}

\subsubsection{Multiple Plant Screening Analysis}

The Screening Interface is designed for efficient analysis of multiple plants simultaneously. This framework comprises three specialized analysis modules, each optimized for specific aspects of plant development while maintaining multiple plants high-throughput processing capabilities, with their corresponding metrics summarized in Table \ref{tab:metrics} under Multiple Plant Analysis.

The germination analysis module implements validated approaches from previous germination analysis systems \cite{ohlsson2024spiro} to monitor seed morphology changes and detect emergence events. The module employs the Four-Parameter Hill Function to model germination progression:

\[ G(t) = G_0 + \frac{G_{max} \cdot t^n}{t_{50}^n + t^n}, \]

\noindent where $G_0$ represents the base germination level, $G_{max}$ is the maximum germination percentage, $n$ denotes the steepness parameter, and $t_{50}$ represents the time to 50\% germination. The Time of Maximum Germination Rate (TMGR) is calculated as:

\[ TMGR = t_{50} \cdot (\frac{n-1}{n+1})^{1/n}. \]

The hypocotyl analysis module incorporates validation steps for reliable measurement in multi-plant scenarios. The system automatically detects physiologically impossible growth rates and artifacts that can occur when different plants touch and their segmentations combine, \textcolor{black}{incorporating biological constraints by forcing non-decreasing length measurements. To ensure full transparency, the platform preserves all raw, unfiltered measurements alongside the processed results. These files are exported in standard formats, allowing researchers to apply custom validation logic or modify the underlying code for specialized needs.}

The plant analysis module provides rapid quantification of basic growth parameters through efficient skeletonization techniques. While not as detailed as the graph-based analysis of the Standard Interface, this module extracts fundamental measurements such as main root length, root area, and full plant area, enabling effective high-throughput screening of general growth patterns.

\subsubsection{Functional Data Analysis of Plant Development}

A major methodological advancement in ChronoRoot 2.0 is the implementation of FPCA for analyzing temporal patterns in plant development. FPCA \cite{fpcasurvey} represents a significant analytical improvement over conventional time-series approaches by treating growth trajectories as continuous functions rather than discrete measurement points.

While traditional plant growth analysis typically relies on point-wise comparisons or summary statistics, which can miss subtle patterns in developmental dynamics, this approach considers the entire growth curve as a functional unit. This enables detection of complex temporal patterns and variations in growth rates that might be overlooked by conventional methods, particularly valuable in plant development studies where the timing and rate of growth can be as biologically relevant as final measurements. For more details and a more graphical explanation of FPCA, we refer the reader to Supplementary Material S1.

FPCA processes temporal measurements through several steps. First, each growth trajectory is converted into a functional data object using monomial basis expansion, providing a continuous representation of the development pattern. The system then performs dimensionality reduction to extract principal component functions that capture the main modes of variation in the data. These functional components are ranked by their explained variance ratio, with typically 2-3 components accounting for over 90\% of the observed variation.

This analysis method can be applied to any temporal measurement extracted by either interface, including root lengths, growth rates, and organ areas. Through quantile-based reconstructions and divergent color palettes, the system provides intuitive visualizations of how components modify developmental trajectories, enabling researchers to detect subtle temporal patterns in growth, identify key time points where developmental trajectories diverge between conditions, and quantify complex growth behaviors through a reduced set of interpretable components.

\subsection{Software Implementation and User Interface}

ChronoRoot 2.0 introduces two dedicated graphical user interfaces developed with Python and PyQt5, replacing the original text-based configuration system. Both interfaces are built upon a shared foundation of scientific computing libraries including NumPy, Pandas, and SciPy for data processing and statistical analysis, OpenCV for image processing, and Matplotlib and Seaborn for visualization.

The Standard Root Phenotyping Interface maintains the core functionality of the original ChronoRoot system while adding modern visualization capabilities. This interface implements a comprehensive analysis pipeline through several interconnected modules (Supplementary Figure \ref{fig:standard_interface_sup}). The main interface provides tools for experimental configuration, ROI-based plant selection, and real-time visualization of segmentation results. Through an intuitive workflow, users can configure analysis parameters, process individual plants, and generate detailed architectural measurements. The analysis capabilities include convex hull analysis, lateral root angle measurements, growth speed evaluation with Fourier analysis, and detailed statistical testing using Mann-Whitney tests at configurable time intervals. Users can specify particular days for detailed reporting and adjust various measurement parameters such as emergence distance for lateral roots. The interface incorporates quality control through visual feedback systems \textcolor{black}{that allow users to inspect the segmentation results prior to plant selection and manually define the root starting position. Once processed, the software generates growth videos overlaid with the resulting graphs and showcases the measurements, enabling researchers to visually validate the tracking performance for each plant. Problematic individuals can then be discarded or re-analyzed before the system proceeds to the automated generation of comprehensive reports and statistical summaries.}

The Screening Interface introduces a streamlined workflow for high-throughput phenotyping experiments  (Supplementary Figure \ref{fig:screening_interface_sup}). The interface guides users through a systematic process from initial calibration to analysis, featuring a dedicated manual calibration tool for precise spatial measurements and an interactive group selection system for defining experimental conditions. Users can define regions of interest corresponding to different treatments or genotypes, and input manual seed counts when needed. The interface implements three specialized analysis modules: germination analysis, hypocotyl development tracking, and basic plant measurements. Real-time visualization tools allow users to monitor segmentation quality, tracking performance, and analysis results as they are generated. 

Both interfaces employ multithreading to maintain responsiveness during computationally intensive operations. Quality control mechanisms are integrated throughout the workflows, enabling users to quickly identify and address potential issues. The system generates automated reports featuring graphical summaries and numerical statistics, making experimental results readily available for analysis and publication. The complete codebase and documentation are freely available through our GitHub repository (detailed in the Data Availability section), enabling reproducibility and further development by the community.

\section{Results}

The performance and capabilities of ChronoRoot 2.0 were evaluated through \textcolor{black}{four} key aspects. First, we assessed the core segmentation capabilities, comparing our nnUNet implementation against the original ChronoRoot system in both accuracy and computational efficiency. Second, we validated the system's multi-class detection capabilities, evaluating its performance in simultaneously identifying and tracking six distinct plant structures \textcolor{black}{across both \textit{Arabidopsis thaliana} and tomato}. \textcolor{black}{Third, we demonstrated the system's multi-species capability through comprehensive evaluation on both species, showcasing robust performance across morphologically distinct plants.} Finally, we demonstrated the software's practical utility through \textcolor{black}{four} comprehensive use cases, showcasing its application in both detailed architectural analysis and high-throughput screening scenarios.

\subsection{Segmentation Performance with nnUNet}

\begin{table}[t!]
\centering
\caption{\textbf{Segmentation performance comparison between original ChronoRoot and ChronoRoot 2.0.} \textcolor{black}{The nnUNet implementation outperforms previous models in both accuracy and processing speed in a separated test set (n=55). All nnUNet configurations achieve higher Dice scores than the original models. While test-time augmentation (TTA) shows no significant impact on segmentation overlap (Dice), it substantially improves boundary precision (Hausdorff), reducing error distances by removing spurious segmentations.}}
\resizebox{\linewidth}{!}{
\begin{tabular}{lccc}
\toprule
\textbf{Model} & \textbf{Dice} & \textbf{Hausdorff (mm)} & \textbf{Processing Time (s)} \\
\midrule
\multicolumn{4}{l}{\textit{Original ChronoRoot}} \\
DSResUNet (Fast) & 0.769 $\pm$ 0.043 & 7.25 $\pm$ 6.87 & $\sim$0.5 \\
Ensemble (Accurate) & 0.772 $\pm$ 0.048 & 7.21 $\pm$ 7.02 & $\sim$4.5 \\
\midrule
\multicolumn{4}{l}{\textit{ChronoRoot 2.0 nnUNet}} \\
Standard & 0.809 $\pm$ 0.041 & 6.41 $\pm$ 5.39 & 2.80 $\pm$ 0.09 \\
Standard (no TTA) & 0.808 $\pm$ 0.042 & 11.08 $\pm$ 13.58 & 0.89 $\pm$ 0.07 \\
Residual & 0.812 $\pm$ 0.038 & 9.07 $\pm$ 9.33 & 5.34 $\pm$ 0.15 \\
Residual (no TTA) & 0.815 $\pm$ 0.032 & 13.10 $\pm$ 14.18 & 1.57 $\pm$ 0.04 \\
\bottomrule
\end{tabular}}
\label{tab:segmentation-comparison}
\end{table}

We first evaluated the segmentation performance of ChronoRoot 2.0's nnUNet implementation against the original ChronoRoot models using their established dataset (consisting of 339 train images and 55 test images) and metrics (Dice coefficient quantifies the overlap between predicted and ground truth segmentations, while the Hausdorff distance measures the maximum boundary error in millimeters), to validate our architectural improvements. This comparison not only validates the new segmentation approach but also demonstrates backward compatibility with the original system's binary segmentation task, ensuring continuity for existing users while providing enhanced capabilities. The nnUNet implementation showed substantial accuracy gains while maintaining practical processing speeds for high-throughput applications (Table \ref{tab:segmentation-comparison}).

The original ChronoRoot system offered two operational modes: a rapid DSResUNet implementation (~0.5 seconds/image) and a more accurate but slower ensemble method (~4.5 seconds/image). While the fast method enabled high-throughput processing, its accuracy (Dice coefficient: 0.769) limited its utility for detailed architectural studies. The ensemble approach achieved marginally better accuracy (Dice: 0.772) but at a significant computational cost.

Regarding ChronoRoot 2.0, \textcolor{black}{we trained two different nnUNet architectural configurations: the standard convolutional architecture and a novel incorporation of a residual encoder architecture \cite{nnunetrevisited}. Our implementation allows users to activate or deactivate test-time augmentation (TTA) at inference time, providing a flexible trade-off between processing speed and segmentation quality. All nnUNet configurations substantially outperformed the original ChronoRoot models, achieving Dice coefficients above 0.808 while maintaining practical processing speeds. Disabling TTA reduces inference time by approximately 3-fold (from 2.80 to 0.89 seconds for standard architecture, and from 5.34 to 1.57 seconds for residual), enabling high-throughput processing. Importantly, test-time augmentation showed divergent effects on the two evaluation metrics: TTA had no significant impact on Dice coefficients, comparing architectures with and without TTA revealed nearly identical overlap performance, yet dramatically improved boundary precision as measured by Hausdorff distance, reducing error distances by 40-45\%. This improvement stems from TTA's ability to remove spurious segmentations through prediction averaging, which primarily affects boundary outliers rather than overall segmentation overlap.}

All training and inference time evaluations were conducted on a standard workstation equipped with an Intel(R) Core(TM) i7-8700 CPU, 64 GB RAM, and an NVIDIA Titan X GPU. 

\begin{table}[t!]
\centering
\label{tab:arabidopsis-summary-metrics}
\caption{\textbf{Multi-class segmentation performance on the \textit{Arabidopsis thaliana} dataset.} \textcolor{black}{All model configurations achieve similar results (n=176). Notably, the fast variants provide a significant reduction in processing time with a minor loss in segmentation performance or structural correctness.}}
\resizebox{\linewidth}{!}{
\begin{tabular}{lcccccc}
\toprule
Model & Dice & HD & Cp & Cr & Time (s) \\
\midrule
Standard & 0.763±0.196 & 8.519±12.500 & 0.934±0.074 & 0.937±0.106 & 2.821 \\
Standard (Fast) & 0.758±0.198 & 9.199±13.312 & 0.929±0.082 & 0.936±0.110 & 0.972 \\
Residual & 0.764±0.193 & 8.415±11.983 & 0.930±0.104 & 0.937±0.109 & 5.300 \\
Residual (Fast) & 0.763±0.189 & 8.743±12.324 & 0.926±0.109 & 0.935±0.113 & 1.630 \\
\bottomrule
\end{tabular}}
\label{tab:arabidopsis-summary-metrics}
\end{table}

\subsubsection{Multi-Class Segmentation Performance}

Building upon these improvements in binary segmentation, we evaluated the nnUNet's performance in discriminating among six distinct plant structures. This multi-class capability represents a significant advancement over the original system, enabling tracking of multiple plant organs throughout development. The dataset was partitioned, within each of the three major experimental categories (etiolation, germination, and plant root analysis), into training (70\%), validation (10\%), and test (20\%) sets following a video-based splitting strategy to prevent data leakage.

Beyond standard segmentation overlap metrics, successful root system analysis depends critically on preserving key morphological traits. We therefore evaluated the skeletonized root segmentations using completeness and correctness metrics, which directly assess structural fidelity \cite{youssef2015evaluation}. Completeness measures the extent to which the extracted skeleton retains the original root structure, with higher values indicating fewer missing segments. Correctness evaluates the presence of extraneous or spurious branches in the extracted skeleton, with high values indicating that the segmentation accurately follows the true root architecture without introducing artifacts.

\textcolor{black}{Table \ref{tab:arabidopsis-summary-metrics} presents the overall performance averaged across all plant structures, and the completeness and correctness calculated for the complete root, for each model configuration. All variants achieved similar Dice coefficients, with the standard and residual architectures showing no significant differences between them, but both significantly outperforming their respective fast (non-TTA) variants according to Wilcoxon Pair Ranked Test. Processing times ranged from ~1 to ~5 seconds per image, with the fast variants providing approximately 3-fold speedup. Detailed per plant organ values are shown in Supplementary Table \ref{tab:arabidopsis-detailed-metrics}.}

\subsubsection{Across species generalization: Tomato.}

\textcolor{black}{To evaluate the generalizability of our approach to other plant species, we trained and tested nnUNet models on a tomato dataset. The data was partitioned by experimental setup, with one complete acquisition (24 plates) reserved for testing, resulting in 299 training images and 181 test images. Note that in the tomato dataset, leaves and petioles were annotated as a single combined aerial part class, reflecting species-specific morphological differences from \textit{Arabidopsis thaliana}.}

\textcolor{black}{We evaluated two training strategies: (1) models trained exclusively on tomato data, and (2) models trained on combined tomato and \textit{Arabidopsis thaliana} datasets. For the combined training approach, \textit{Arabidopsis} annotations were preprocessed to match the tomato class structure by merging leaf and petiole classes into a single aerial part category. All four model configurations (Standard, Standard Fast, Residual, and Residual Fast) were evaluated under both training regimes.}

\textcolor{black}{The results (Table \ref{tab:tomato-summary-metrics}) reveal several important findings. First, residual architectures consistently outperformed standard architectures across all metrics, with particularly notable improvements in Hausdorff distance and correctness measures. Second, incorporating \textit{Arabidopsis} training data significantly enhanced performance across all model configurations, with the combined training strategy yielding the best results. These findings demonstrate both the transferability of knowledge across plant species and the value of diverse training data for robust segmentation performance. Detailed per plant organ values are shown in Supplementary Table \ref{tab:tomato-detailed}.}

\begin{table}[t!]
\centering
\caption{\textbf{Cross-species generalization and training strategy evaluation for the tomato dataset.} \textcolor{black}{Residual architectures show better results in Hausdorff Distance (HD) and Correctness (Cr), in the separated test set (n=181). The multi-species training strategy (Both) consistently outperforms training only on tomato data, showing that data diversity improves results across different morphologies.}}
\label{tab:tomato-summary-metrics}
\resizebox{\linewidth}{!}{
\begin{tabular}{lcccccc}
\toprule
Training & Configuration & Dice & HD & Cp & Cr & Time (s) \\
\midrule
Tomato & Standard &  0.815±0.218 & 19.793±20.590 & 0.920±0.133 & 0.779±0.230 & 2.198 \\
Tomato & Standard (Fast) & 0.801±0.220 & 24.357±22.973 & 0.910±0.157 & 0.733±0.253 & 0.813  \\
Tomato & Residual & 0.843±0.200 & 15.430±18.900 & 0.908±0.153 & 0.868±0.189 & 5.290 \\
Tomato & Residual (Fast) & 0.829±0.206 & 17.295±19.492 & 0.904±0.159 & 0.853±0.197 & 1.586 \\
Both & Standard & 0.828±0.212 & 19.803±20.335 & 0.914±0.141 & 0.816±0.223 & 2.199 \\
Both & Standard (Fast) &  0.822±0.214 & 19.722±19.897 & 0.908±0.156 & 0.791±0.232 & 0.768 \\
Both & Residual & 0.863±0.201 & 11.089±16.640 & 0.916±0.138 & 0.896±0.177 & 5.310 \\
Both & Residual (Fast) & 0.858±0.195 & 11.553±16.174 & 0.905±0.165 & 0.899±0.172 & 1.557  \\
\bottomrule
\end{tabular}
}
\end{table}

\subsection{Demonstration of Software Capabilities Through Use Cases}

To validate ChronoRoot 2.0's practical utility across diverse experimental scenarios, we implemented three use cases. The first one examines root system architecture under long day and continuous light condition following the original publication \cite{gaggion2021chronoroot}, while the second analyzes published data from the transcription factor gene \textit{NF-YA10} over-expressing plants \cite{barrios2025transcription}. The final use case demonstrates the high-throughput screening capabilities of the system, on an etiolation experiment. 
Importantly, all figure subpanels presented in these use cases are direct outputs from ChronoRoot 2.0 \textcolor{black}{and serve as representative examples of the automated reports generated when users analyze their own data or the provided demo datasets}. The only modification \textcolor{black}{to these outputs} is the addition of asterisks to indicate statistical significance: \textcolor{black}{\textasteriskcentered{} for p<0.05 and \textasteriskcentered{}\textasteriskcentered{} for p<0.001}. \textcolor{black}{While the figures prioritize visual clarity for phenotype comparison, the exact} statistical analysis values\textcolor{black}{—including p-values, means, and standard deviations—}are provided by the software as accompanying text files \textcolor{black}{within the output folders}, requiring no additional analysis beyond what the software automatically generates.

\begin{figure*}[t!]
\centering
\includegraphics[width=\linewidth]{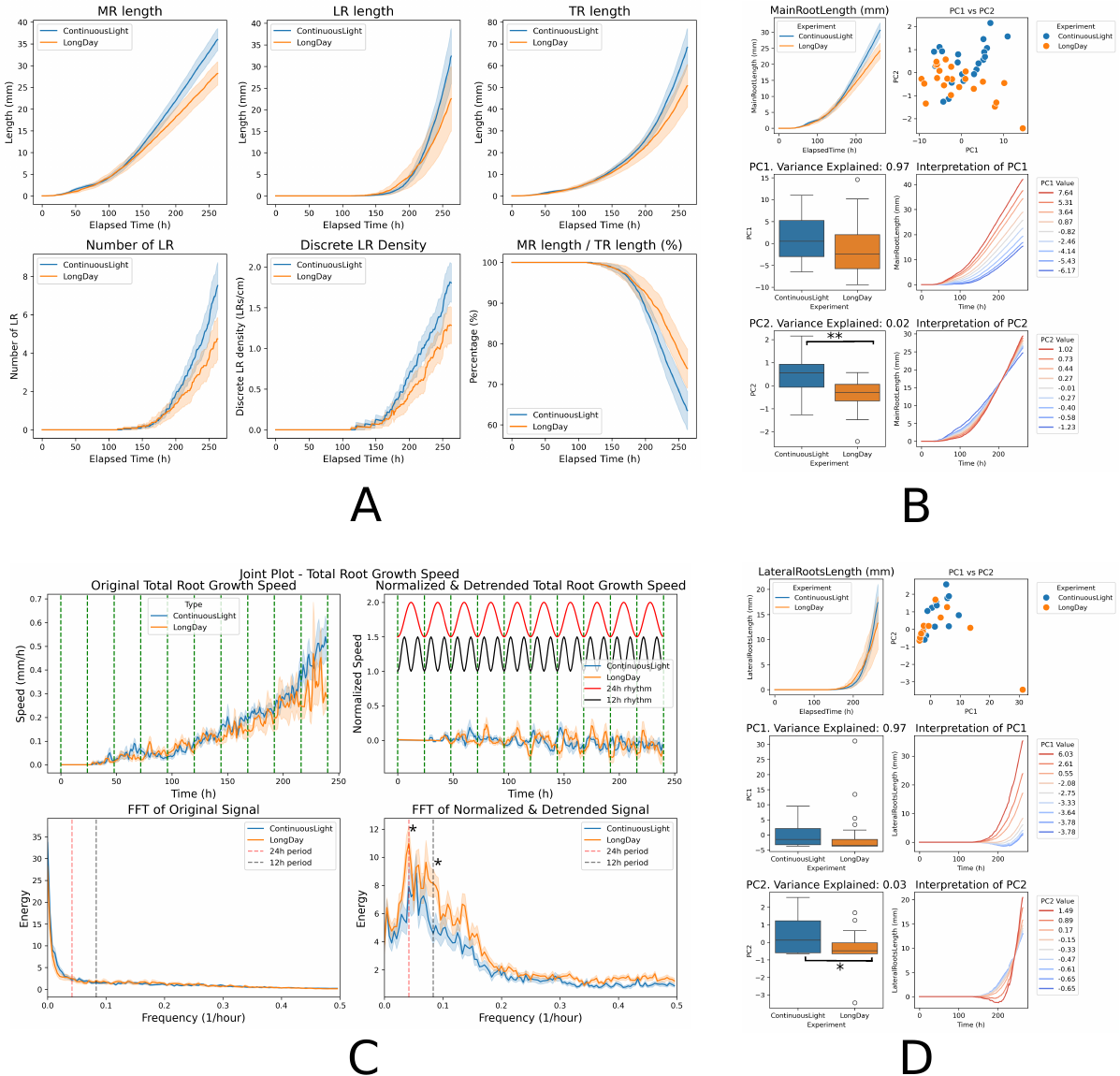}
\caption{\textbf{Use Case 1 -  \textcolor{black}{\textit{Arabidopsis thaliana}} Root system architecture dynamics under different light conditions.} Comparison of long-day (16h/8h, blue, n=23) versus continuous light (24h, orange, n=21) shows divergent growth patterns. A - All basic Architectural RSA parameters, B - FPCA analysis of Main Root Length, \textcolor{black}{significative differences found in PC2 (p-value<0.001)}, \textcolor{black}{C - Fourier transform of Total Root Growth Rate (significative differences found at both 24h and 12h periods, p-value<0.05)}, D - FPCA analysis of Lateral Root Length, \textcolor{black}{significative differences found in PC2 (p-value<0.05)}. Error bands: standard error.}
\label{fig:use_case_1}
\end{figure*}

\subsubsection{Use Case 1 - Temporal Dynamics of Root System Architecture: Replication and Extension of ChronoRoot Findings with Fourier and FPCA}

\noindent \textbf{Plant materials:} \textit{Arabidopsis thaliana} ecotype Col-0 seeds were surface sterilized and stratified at 4°C for 2d before being grown under long day conditions (16h light, $140 \mu E m{}^{-2}s{}^{-1}$/ 8h dark), or continuous light (24h light, $140 \mu E m{}^{-2}s{}^{-1}$) at 22°C, on half-strength Murashige and Skoog media (1/2 MS) (Duchefa, Netherlands) with 0.8\% plant agar (Duchefa, Netherlands). Four seeds were used per plate. 

Root system architecture exhibits complex temporal dynamics that can reveal fundamental aspects of plant adaptation to environmental conditions. Building upon the findings reported in \cite{gaggion2021chronoroot}, we investigated how different light regimes influence root development patterns, leveraging our enhanced analytical capabilities to uncover subtle temporal variations in growth dynamics.

To validate and extend the findings from ChronoRoot, we replicated its analysis pipeline and incorporated FPCA to further dissect the temporal dynamics of RSA. First, we computed conventional RSA metrics, including main root length, lateral root length, total root length, lateral root density, and the proportion of the main root relative to total root length (Figure \ref{fig:use_case_1}-A). We then explored root growth dynamics by applying FPCA to the temporal evolution of root length.

The first functional principal component (PC1) captured the primary growth trajectory of roots (Figure \ref{fig:use_case_1}-B), revealing differences between photoperiod conditions. The second (PC2) showed distinct divergence between long-day and continuous-light conditions, indicating temporal shifts in growth patterns. Following the ChronoRoot methodology, we also analyzed root elongation rates through Fourier Transform (Figure \ref{fig:use_case_1}-C) to detect underlying oscillatory patterns, identifying both circadian (24-hour) and ultradian (12-hour) rhythms under long-day conditions, which were disrupted under continuous light. Similar analysis of lateral root length (Figure \ref{fig:use_case_1}-D) showed comparable patterns.

\subsubsection{Use Case 2 - Complete RSA Characterization of different \textit{Arabidopsis thaliana} genotypes: Area covered, lateral root angles and tip angle decay over time}

\noindent \textbf{Plant materials:} All plants used in this study are in Columbia-0 background. pNF-YA10:GFP-NF-YA10miRres (NF-YA10miRres) stable lines were obtained by transforming \textit{Arabidopsis} plants with a construct bearing \textcolor{black}{2000 bp} region upstream of the start codon of NF-YA10 amplified from genomic DNA (promoter region) and the coding sequence (CDS) of \textit{NF-YA10} without miRNA cleavage site amplified from cDNA, thus resisting  miR169-mediated post-transcriptional silencing of \textit{NF-YA10} mRNA. More details were published at \cite{barrios2025transcription}.

\begin{figure*}[!t]
\centering
\includegraphics[width=1\linewidth]{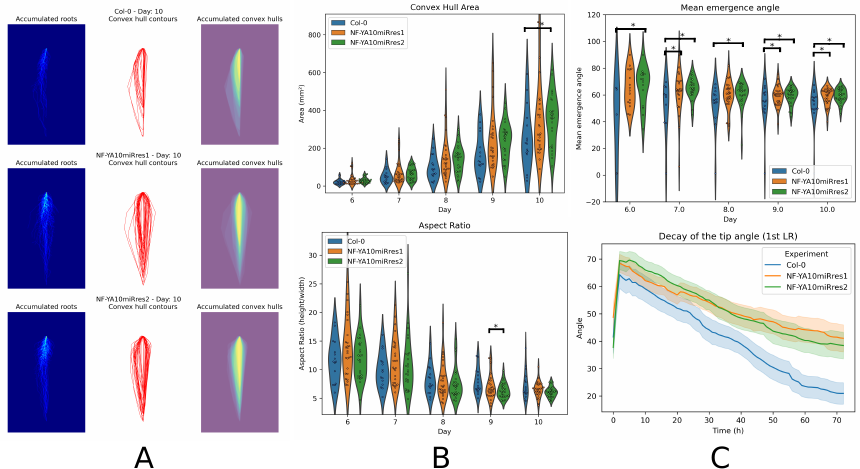}
\caption{\textbf{Use Case 2 - Altered root architecture in \textcolor{black}{\textit{Arabidopsis thaliana}} NF-YA10miRres plants.} A - Qualitative analysis of convex hull showing root system coverage differences between genotypes. B - Quantitative analysis of convex hull metrics (area and aspect ratio) between NF-YA10miRres plants \textcolor{black}{(1 in orange, n=26), 2 in green, n=28)} and Col0 controls \textcolor{black}{(blue, n=16)}. 
C - Quantitative analysis of average emergence angle and base-tip angle for the first lateral root, demonstrating consistently wider angles in NF-YA10miRres plants compared to Col0. Error bars: standard error.}
\label{fig:use_case_2}
\end{figure*}

ChronoRoot 2.0's enhanced analytical capabilities revealed distinct architectural patterns between NF-YA10miRres and wild-type Col0 plants. Using the convex hull analysis (Figure \ref{fig:use_case_2}-A and B), we quantified the overall root system distribution. Qualitative visualization (Figure \ref{fig:use_case_2}-A) and quantitative metrics (Figure \ref{fig:use_case_2}-B) showed that NF-YA10miRres plants developed significantly larger convex hull areas, indicating broader root system coverage. Moreover, these plants exhibited higher aspect ratios (height/width), suggesting that lateral roots grew at wider angles from the main root axis rather than clustering around it.

The novel angle measurement capabilities provided detailed insights into these architectural differences. During normal root development, lateral roots typically exhibit gravitropic responses, gradually bending downward after emergence - a phenomenon we term 'angle decay'. 
Temporal analysis of these lateral root angles (Figure \ref{fig:use_case_2}-C) showed that NF-YA10miRres plants consistently maintained larger angles compared to wild-type, indicating an altered gravitropic response. The base-tip angle difference progressively increased, reaching a 20° differential after three days of growth. The emergence angles showed similar trends, becoming significantly different from wild-type by day 9. This temporal progression of angular differences suggests that the transcription factor NF-YA10 plays a role in regulating both the initial trajectory and subsequent gravitropic responses of lateral roots.

\subsubsection{Use Case 3 - High-throughput Analysis of Etiolation in \textit{Arabidopsis thaliana} seedlings}

\begin{figure*}[t!]
    \centering
    \includegraphics[width=0.78\linewidth]{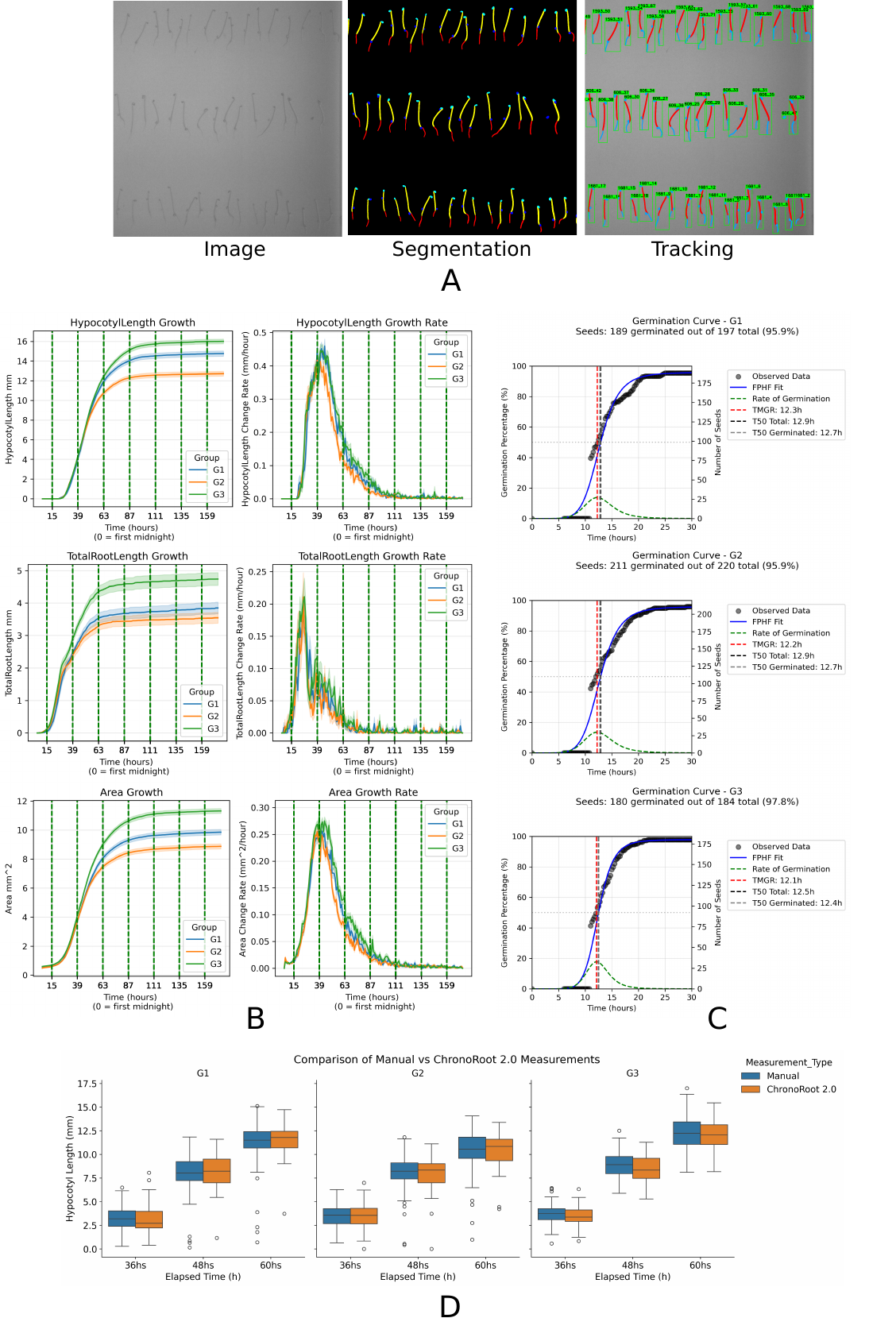}
    \caption{\textbf{Use Case 3 - High-throughput analysis of  \textcolor{black}{\textit{Arabidopsis thaliana}} seedling etiolation.} A - Representative infrared images showing temporal progression of etiolated seedling development.
    B - Hypocotyl, root and total area measurements with their corresponding growth rates (G1, blue, n=189. G2, orange, n=211; G3, green, n=180). C - Germination curves showed T50 at approximately 13 hs after light stimulation with no significant differences between the analyzed genotypes. D - Comparison of Manual (blue) and automatic (orange) hypocotyl length determinations at 36, 48 and 60 hours after light stimulation (no significant difference observed). Error bars: standard error.}
    \label{fig:use_case_3}
\end{figure*}

\noindent \textbf{Plant materials:} Lines G1, G2, and G3 were in the Columbia-0 (Col0) background. Seeds were surface sterilized and sown on MS medium supplemented with 1\% agar in 120mm-side square petri dishes. To maximize the experimental throughput, up to 100 seeds were placed in each plate in a grid pattern, with 3 rows of 33 seeds.

To demonstrate ChronoRoot 2.0's capabilities for high-throughput phenotyping under specific growth conditions, we conducted an etiolation study across three genotypes. After brief light exposure for germination synchronization, plants were grown in complete darkness for 5 days, with automated infrared imaging every 15 minutes to track development without light interference.

The system's enhanced segmentation algorithm successfully distinguished between hypocotyl and root tissues, enabling precise quantification of both structures' growth dynamics. It was also able to identify smaller structures like the cotyledons (embryonic leaves) and the seed coverage (Figure \ref{fig:use_case_3}-A). All genotypes exhibited the characteristic etiolation response, with dramatic hypocotyl elongation as seedlings searched for light. Hypocotyl length measurements revealed significant differences in elongation between the three genotypes, with Genotype G3 exhibiting the largest hypocotyl, followed by G1 and then G2 (Figure \ref{fig:use_case_3}-B). The same differences can be appreciated in the growth rate curves, showing that G3 had more sustained growth, followed by G1 and then G2 (Figure \ref{fig:use_case_3}-B). Analysis of root system development showed the same temporal pattern as the hypocotyl: growth during the first 3 days followed by complete growth cessation after day 4, marking the exhaustion of seed reserves under dark conditions. This pattern was clearly visible in both main root length progression and area coverage metrics (Figure \ref{fig:use_case_3}-B).

All genotypes showed the germination of 50\% of the seedlings between 13.04 and 13.78 h post light stimulation and did not differ significantly (Figure \ref{fig:use_case_3}-C). This implied that hypocotyl length differences are due to higher growth rate or sustained growth rather than differences on germination time.

Comparison with manual hypocotyl measurements performed on the same dataset showed no statistically significant differences, demonstrating the robustness of the segmentation and measuring process (Figure \ref{fig:use_case_3}-D). We compared measures at 36, 48 and 60 hours for the three genotypes.

To further characterize developmental patterns across genotypes, we applied FPCA to the growth trajectories (Supplementary Figure \ref{fig:use_case_3_sup}). This analysis revealed that over 97\% of the variance in hypocotyl length, root length, and total plant area could be explained by just two principal components. The first component (PC1) primarily captured differences in final plant size/length, while the second component (PC2) represented temporal shifts in the growth pattern, similarly to the PC2 in Use Case 1. FPCA scores confirmed the genotype differences observed in the direct measurements, with G3 showing significantly higher PC1 scores for hypocotyl elongation and area development, followed by G1 and then G2, as expected.

\subsubsection{Use Case 4 - Multi-species Capability: Tomato Analysis}

\noindent \textbf{Plant materials:} \textcolor{black}{Tomato seeds of cultivar M82 were surface sterilized and sown on MS medium under standard conditions. To accommodate the larger size of tomato seedlings compared to \textit{Arabidopsis}, plant density was reduced to two seeds per plate.}

\begin{figure*}
    \centering
    \includegraphics[width=0.75\linewidth]{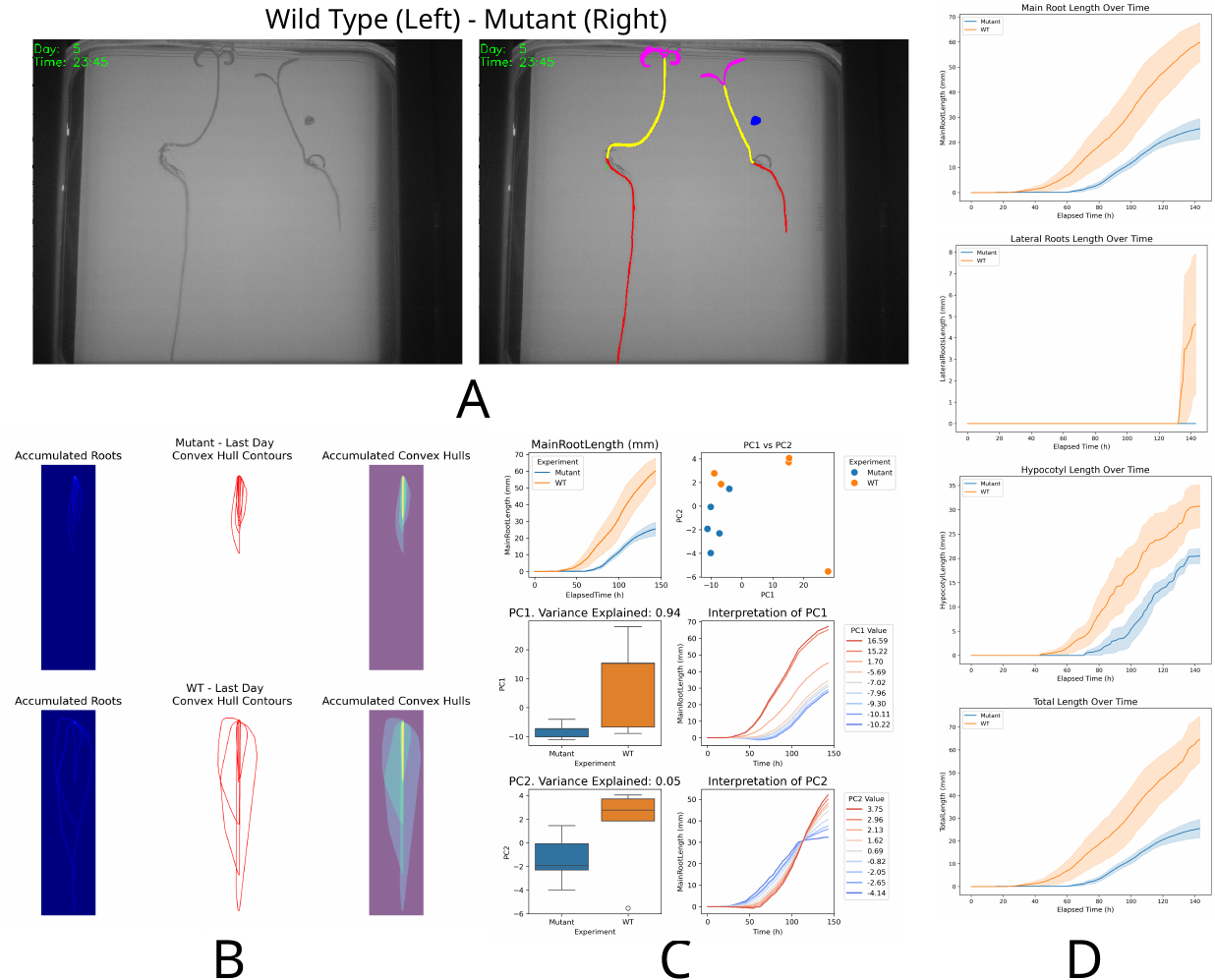}
    \caption{\textcolor{black}{\textbf{Use Case 4 - ChronoRoot 2.0 multi-species analysis on tomato seedlings.}
    A - Representative time-lapse images of wild type (WT, left, n=5) and mutant (right, n=5) tomato seedlings, with nnUNet-based segmentation and organ tracking overlays.
    B - Spatial characterization of root system architecture showing accumulated root traces, last-day convex hull contours, and occupied growth area for mutant (top) and WT (bottom).
    C - Multivariate analysis of extracted traits: main root length dynamics, PCA variance explanation, boxplots for PC1 and PC2, and temporal interpretation of principal components.
    D - Organ-specific temporal growth curves for main root, lateral roots, hypocotyl, and total length, comparing WT and mutant seedlings.}}
    \label{fig:use_case_4}
\end{figure*}

\textcolor{black}{
To further demonstrate the species-agnostic design of ChronoRoot~2.0, we applied the complete analysis pipeline to tomato seedlings, which present larger organs, thicker roots, and increased curvature compared to \textit{Arabidopsis}. The nnUNet-based segmentation accurately identified and tracked main roots, lateral roots, and hypocotyls over time in both wild type and mutant plants (Figure~\ref{fig:use_case_4}-A), enabling the extraction of standard architectural and temporal traits without parameter tuning.}

\textcolor{black}{Quantitative analysis revealed marked differences between the two genotypes. Spatial descriptors such as accumulated root traces and convex hulls showed that mutant seedlings explored a substantially smaller area than wild type plants (Figure~\ref{fig:use_case_4}-B). Temporal measurements of main root, lateral root, hypocotyl, and total length highlighted an early and persistent reduction in growth in the mutant condition (Figure~\ref{fig:use_case_4}-D). Principal component analysis of the extracted traits captured most of the variance with the first two components and clearly separated wild type and mutant populations, reflecting differences in overall growth magnitude and temporal progression (Figure~\ref{fig:use_case_4}-C).}

\section{Discussion}

ChronoRoot 2.0 provides plant biologists with an integrated solution for analyzing root system development across multiple experimental scales, integrated with additional parameters of the seedling aerial organs. While the artificial nature of 2D growth systems on petri dishes represents an inherent limitation in root architecture studies, our results demonstrate how enhanced measurement capabilities can reveal meaningful biological patterns even within these constraints.

The multi-class segmentation approach addresses a significant challenge in developmental studies by enabling simultaneous analysis of multiple plant structures. Although root growth on agar plates differs from soil conditions, the ability to precisely track both below and above-ground organs provides valuable insights into developmental coordination. The etiolation response study demonstrates how this capability can reveal resource allocation patterns during early development, with the simultaneous tracking of hypocotyl elongation and root growth providing a more complete understanding of seedling responses to dark conditions. \textcolor{black}{To demonstrate the system's adaptability beyond \textit{Arabidopsis thaliana}, we incorporated a tomato (\textit{Solanum lycopersicum}) dataset. This addition validates that the self-configuring nnUNet core can be effectively retrained to handle the more robust and diverse morphologies of crop species. While applying the system to plants with fundamentally different architectures would require new annotated training data, this framework ensures that the adaptation process remains accessible to researchers without deep machine learning expertise.}

\textcolor{black}{While the software architecture is fundamentally modality-agnostic and offers a modular pathway for adaptation to other 2D imaging platforms (such as SPIRO \cite{ohlsson2024spiro}), the platform’s robustness is anchored in its integration with our custom hardware. By coupling the software with an open-source, affordable, and easily assembled hardware unit, we ensure high data quality and stable temporal resolution (15-minute intervals) without presenting a significant financial barrier to adoption.}

\textcolor{black}{To maximize the utility of these segmentation capabilities within the spatial limits of 12x12 cm plates, we implemented a dual-interface system designed to handle specific experimental constraints. The \textit{Standard Interface} is designed for high-precision architectural tracking of normally 4-6 plants per plate. These experiments are typically limited to 10-14 days for \textit{Arabidopsis} or 5-7 days for tomato, concluding when the main root reaches the plate bottom or grows along the surface invisible to imaging. Conversely, the \textit{Screening Interface} accommodates up to 100 plants per plate but limits analysis to the early developmental window (typically 3-5 days). This mode allows for massive data collection before plant crowding and physical contact prevent reliable segmentation.}

The automated angle measurements introduce new possibilities for quantifying gravitropic responses in standardized conditions. While plate-based growth systems impose spatial constraints on root architecture, the precise measurement of emergence angles and their temporal evolution, as demonstrated in the NF-YA10miRres vs. Col0 analysis, enables systematic study of gravitropic regulation. These measurements provide a standardized framework for comparing gravitropic responses across genotypes and conditions, even within the limitations of 2D growth systems.

The temporal analysis capabilities represent a particular strength for understanding dynamic developmental processes. The identification of distinct growth rhythms under different light conditions demonstrates how high-resolution temporal data can reveal patterns that might be missed by endpoint measurements. While circadian patterns in artificial growth conditions may differ from natural environments, the ability to detect and quantify these rhythms provides valuable insights into the temporal organization of plant development. Moreover, the incorporation of FPCA-based analysis facilitates the interpretation of complex temporal signals by reducing their dimensionality, providing a novel and easily-explainable way to quantify dynamic growth patterns. 

The analysis frameworks implemented in ChronoRoot 2.0 open new possibilities for understanding plant development, even within the constraints of traditional growth systems. The ability to automatically quantify subtle architectural differences and temporal patterns enables systematic comparison of developmental responses across genotypes and conditions. These capabilities are particularly valuable for studies investigating the genetic and environmental regulation of plant development, where precise quantification of phenotypic differences is essential. Furthermore, high-throughput comprehensive phenotyping emerges as a powerful tool for genome-wide association studies and the identification of key genes participating in plant development.

ChronoRoot 2.0's release as an open-source platform represents our commitment to accessible, community-driven plant phenotyping tools. While the current implementation provides robust capabilities for analyzing plate-based growth experiments, the modular architecture and comprehensive documentation enable researchers to adapt and extend the system for their specific needs. By releasing both the software and hardware specifications openly, we aim to foster a collaborative community where researchers can share improvements, analytical modules, and experimental protocols. We hope that this approach to open science will not only ensure transparency and reproducibility but also allow the system to evolve alongside the changing needs of the plant biology community.

\section{Code and Data Availability}

The data supporting the findings of this study consists of three main components:

\begin{itemize}
    \item The complete source code of ChronoRoot 2.0, including the implementation of all analysis methods described in this paper, is freely available under the GNU General Public License v3.0 at \url{https://github.com/ChronoRoot/ChronoRoot2}. This repository contains the full software implementation and comprehensive documentation to set up the system and utilizing the software.
    
        \begin{itemize}
        \item Project name: ChronoRoot 2.0
        \item Project home page: \url{https://chronoroot.github.io}
        \item Main Source Code repository: \url{https://github.com/ChronoRoot/ChronoRoot2}
        \item Operating system(s): Platform independent
        \item Programming language: Python
        \item Other requirements: \textcolor{black}{Conda, Apptainer, or} Docker
        \item License: GNU GPL 3.0
        \end{itemize}
        
    \item \textcolor{black}{The annotated image dataset used for training and validation contains 911 infrared images of \textit{Arabidopsis thaliana} seedlings and 480 images of tomato with expert annotations for multiclass segmentation. This dataset is publicly available without restrictions at \url{https://huggingface.co/datasets/ngaggion/ChronoRoot2}. The dataset includes both raw images and their corresponding multi-class segmentation masks in .nii.gz format, as directly generated in the manual annotations made by our biologists. Scripts to convert to nnUNet's standardized structure for 2D images are also provided within the GitHub repo, generating the correct splits to avoid mixing videos in training, validation and test partitions.}

    \item \textcolor{black}{To facilitate reproducibility and allow users to familiarize themselves with the different analysis modules, we provide four complete demo datasets covering the scenarios presented in this paper. These datasets are available within the apptaniner and Docker images, via the ChronoRoot website, and as Supplementary Material to this manuscript:
    \begin{enumerate}
        \item \textbf{Detailed Root Analysis:} A video for RSA characterization of individual \textit{Arabidopsis} plants.
        \item \textbf{Germination Screening:} A video containing hundreds of seeds to test the germination analysis module.
        \item \textbf{Etiolation Screening:} A video of seedlings grown in darkness for testing hypocotyl growth rates.
        \item \textbf{Tomato Comparison:} A pair of videos of tomato illustrating cross-species capability.
    \end{enumerate}
    }

    \item To facilitate deployment and ensure reproducibility across different computing environments, we provide a pre-configured Docker image at \url{https://hub.docker.com/r/ngaggion/chronoroot}. This image includes all necessary dependencies and can be used without any installation requirements beyond Docker itself.

\end{itemize}

\section{Declarations}

\subsection{List of abbreviations}

FPCA: Functional principal component analysis; \\
GPU: graphical processing unit; \\
IR: infrared; \\
LR: lateral root; \\
MR: main root; \\
PC: principal component; \\
ROI: region of interest; \\
RSA: root system architecture; \\
RSML: Root System Markup Language; \\
SORT: Simple online realtime tracking algorithm; \\
TMGR: Time of maximum germination rate; \\
TR: total root; \\

\subsection{Ethical Approval}

Not applicable.

\subsection{Consent for publication}

Not applicable.

\subsection{Competing Interests}

The authors declare that they have no competing interests.

\subsection{Funding}

AB, NG, TB, MC and FA benefit from the ECOS‐SUD Exchange Program (no. A20N05) and the IRP LOCOSYM (CNRS). The FA lab is funded by Agencia I+D+i, ICGEB and AXA Research Fund. AB, MC and TB benefited from the support of the French Agence Nationale de la Recherche (Saclay Plant Sciences-SPS, ANR-17-EUR-0007). 

\subsection{Author's Contributions}

\begin{itemize}
    \item N. Gaggion led the project implementation, developed all software components, and wrote the manuscript
    \item R. Bonazzola conceptualized and implemented the Functional Principal Component Analysis methodology
    \item F. Accavallo contributed to the development and enhancement of the ChronoRoot hardware system
    \item F. Ariel, A. Barrios, F. Catulo, T. Blein and \textcolor{black}{N. Boccardo} conducted the biological experiments analyzed in the study
    \item F. S. Rodriguez, F. E. Aballay, F. B. Catulo, M. F. Mammarella, M. F. Legascue, and S. N. Villarreal performed manual image annotation for training the \textcolor{black}{\textcolor{black}{Arabidopsis thaliana}} nnUNet model.
    \item \textcolor{black}{N. Boccardo and L. Santoro performed manual image annotation for training the tomato nnUNet model}.
    \item \textcolor{black}{L. I. Pereyra-Bistrain and M. Benhamed identified the gene involved in root development, generated the tomato mutant and provided the plant material for Use Case 4.}
    \item F. B. Catulo and M. M. Ricardi performed the validation of the hypocotyl length analyzes
    \item F. Ariel, E. Petrillo, M. M. Ricardi, T. Blein and M. Crespi provided oversight of all biological experiments and analyses, and evaluated the quality of the generated reports
    \item E. Ferrante provided oversight of the computational experiments and supervised the software design process
    \item T. Blein, N. Gaggion, M. Crespi, F. Ariel, and E. Ferrante contributed to the original ChronoRoot concept and its evolution to version 2.0, and provided manuscript revision
\end{itemize}

\section{Acknowledgements}

Not applicable.

\section{Supplementary Materials}

The following supplementary materials are provided to ensure reproducibility and facilitate adoption of the ChronoRoot 2.0 system.

\textbf{Text S1} Functional PCA \\
Provides an intuitive explanation of functional principal component analysis (FPCA) for readers without a quantitative background.

\textbf{Figure S1} Functional principal component decomposition of simulated curves. 
Illustrates the fundamental concepts of FPCA through simplified example data.

\textbf{Figure S2} The Standard Root Phenotyping Interface \\
Demonstrates the complete workflow of the detailed architectural analysis pipeline through six tabs: Plant Analysis (main screen), Preview Image, Analysis Overview, Plant Overlay, Generate Report, and Report.

\textbf{Figure S3} The Screening Interface \\
Illustrates the high-throughput analysis workflow through four tabs: Analysis (main screen), Preview Image, Results, and Reports for efficient multi-plant phenotyping.

\textcolor{black}{\textbf{Table S1} Detailed report of segmentation performance across plant organs for \textit{Arabidopsis thaliana}.}

\textcolor{black}{\textbf{Table S2} Detailed report of segmentation performance for tomato}

\textbf{Figure S4} Functional PCA applied to etiolation experiment.

Presents FPCA of hypocotyl length, root length, and area growth curves from Use Case 3, showing mean trajectories by genotype, principal component distributions, and visual interpretations of how PC1 and PC2 modulate developmental patterns across genotypes G1, G2, and G3.

\bibliography{sample}

\newpage

\section*{Supplementary Material}

\subsection*{S1. Functional PCA}
\label{supfpca}
\renewcommand{\thefigure}{S\arabic{figure}}
\setcounter{figure}{0}

\renewcommand{\thetable}{S\arabic{table}}
\setcounter{table}{0}

\begin{figure*}[!ht]
\centering
\includegraphics[width=1\linewidth]{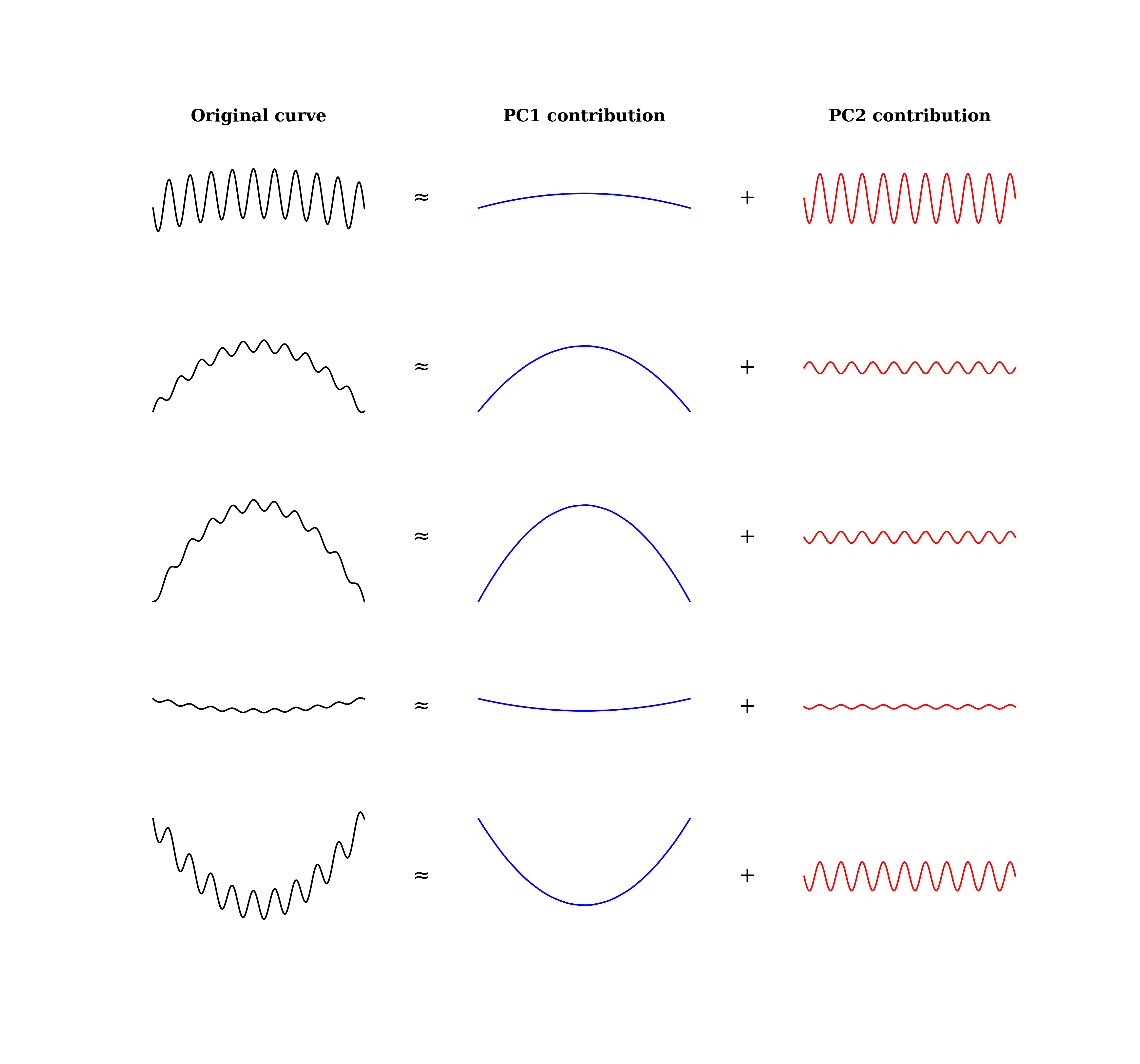}
\caption{
\textbf{Functional principal component decomposition of simulated curves.}
Curves were generated as a linear combination of a quadratic function and a high-frequency sine wave, with coefficients drawn independently from normal distributions:
$ f(x) = a \cdot x^2 + b \cdot \sin(10\pi x)$ with $a \sim \mathcal{N}(0, 1^2)$ and $b \sim \mathcal{N}(0, 0.1^2).$
Each row corresponds to a random instance of such curves. Each curve was centered by subtracting its mean value. Functional PCA was applied to the dataset (using 10,000 randomly sampled curves), and the first two principal components (PCs) were extracted. In each row, the left panel shows the original curve. The middle and right panels show the contributions of the first and second components (PC1 and PC2), respectively. The components are orthogonal and reflect statistically independent sources of variation: the first captures the parabolic shape (due to variation in $a$), while the second captures the oscillatory pattern (variation in $b$).}
\label{fig:fpca_explanation}
\end{figure*}

\begin{figure*}
    \centering
    \includegraphics[width=0.92\linewidth]{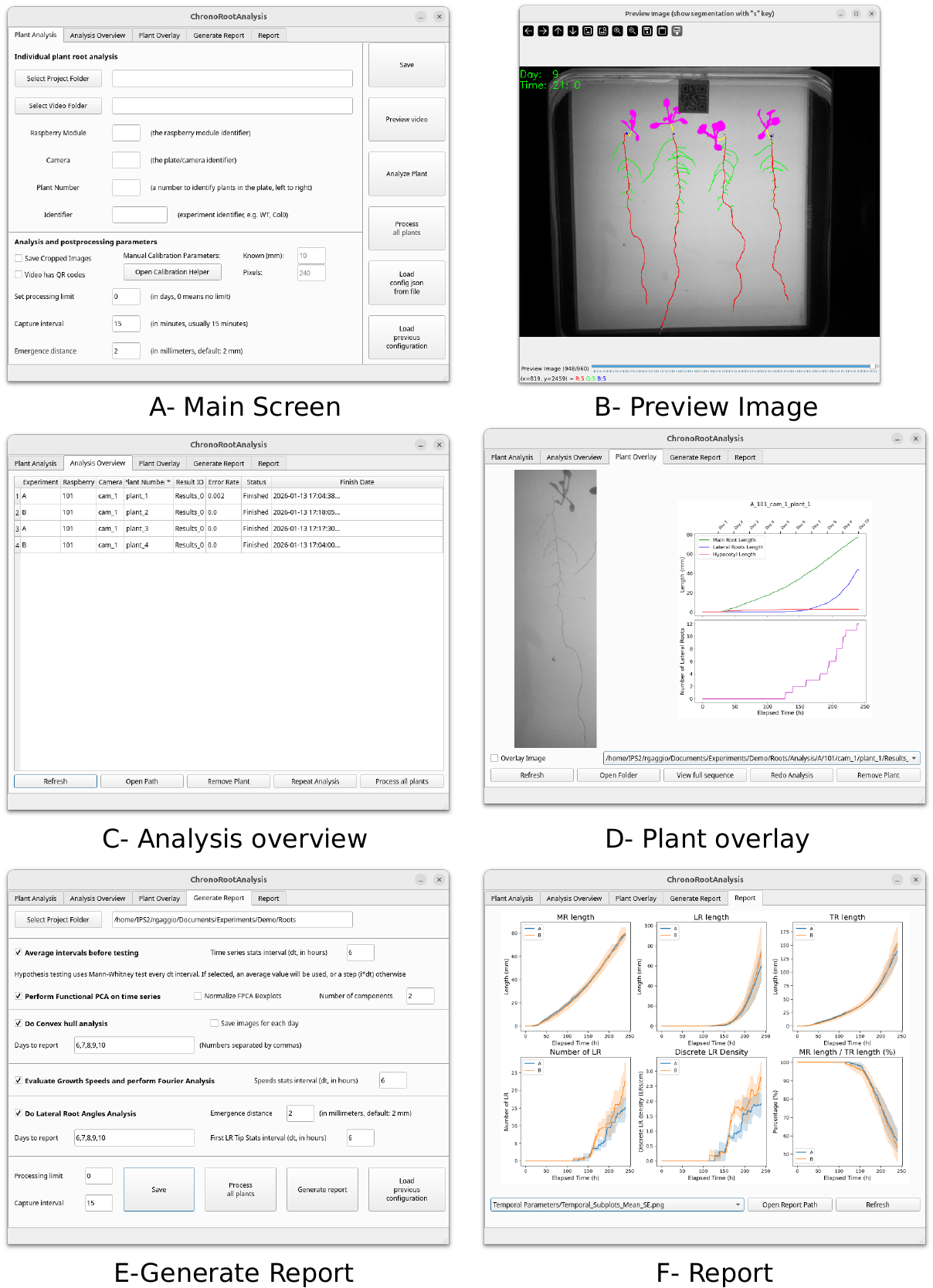}
    \caption{\textbf{The Standard Root Phenotyping Interface:} (A) Main Screen: Plant Analysis tab showing experiment parameters, input/output paths, and processing controls. (B) Preview Image tab with temporal navigation and segmentation toggle for quality assessment. (C) Analysis Overview tab displaying processing completion and error rates. (D) Plant Overlay tab showing individual plant measurements and segmented visualization. (E) Generate Report tab for customizing measurement selection. (F) Report tab displaying finalized architectural analysis results.}
    \label{fig:standard_interface_sup}
\end{figure*}

\begin{figure*}
    \centering
    \includegraphics[width=1.0\linewidth]{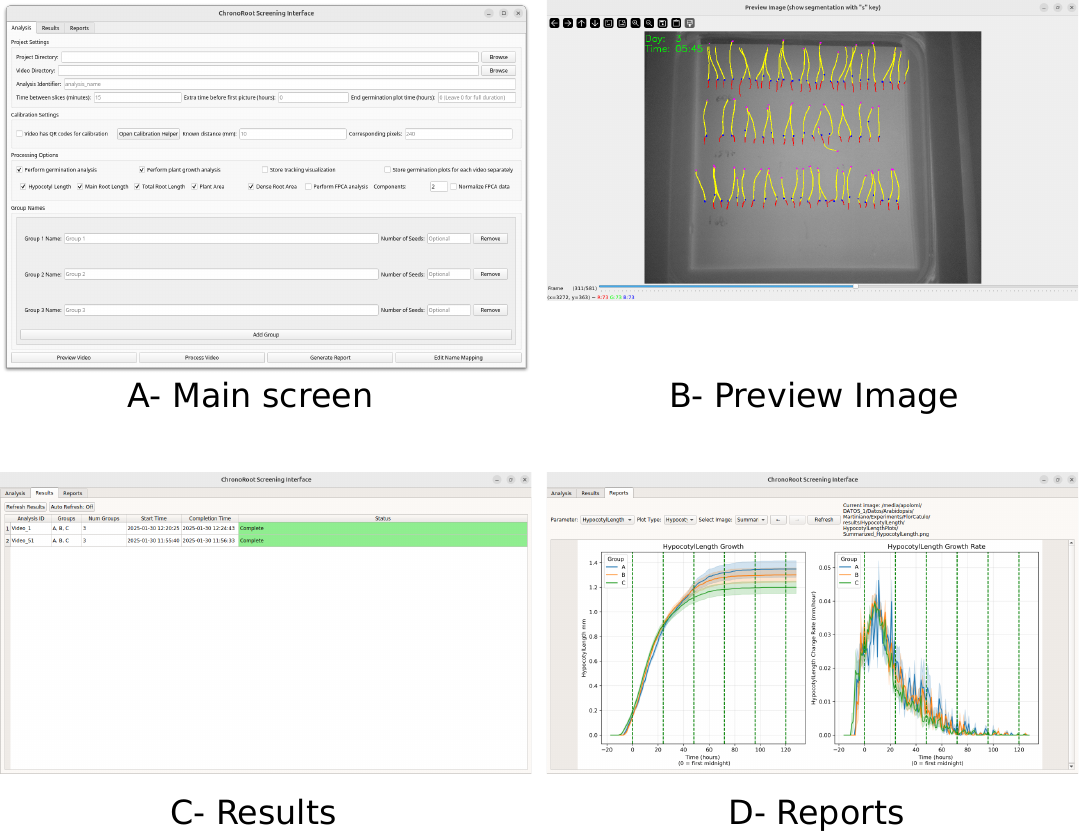}
    \caption{\textbf{The Screening Interface.} (A) Main Screen: Analysis tab with experiment setup, group definition, seed counting, and calibration tools. (B) Preview Image tab providing temporal navigation and segmentation quality assessment. (C) Results tab showing processing status for all plants across experimental groups. (D) Reports tab displaying comparative visualizations and statistical analyses between experimental conditions.}
    \label{fig:screening_interface_sup}
\end{figure*}

\begin{figure*}[!ht]
    \centering
    \includegraphics[width=\linewidth]{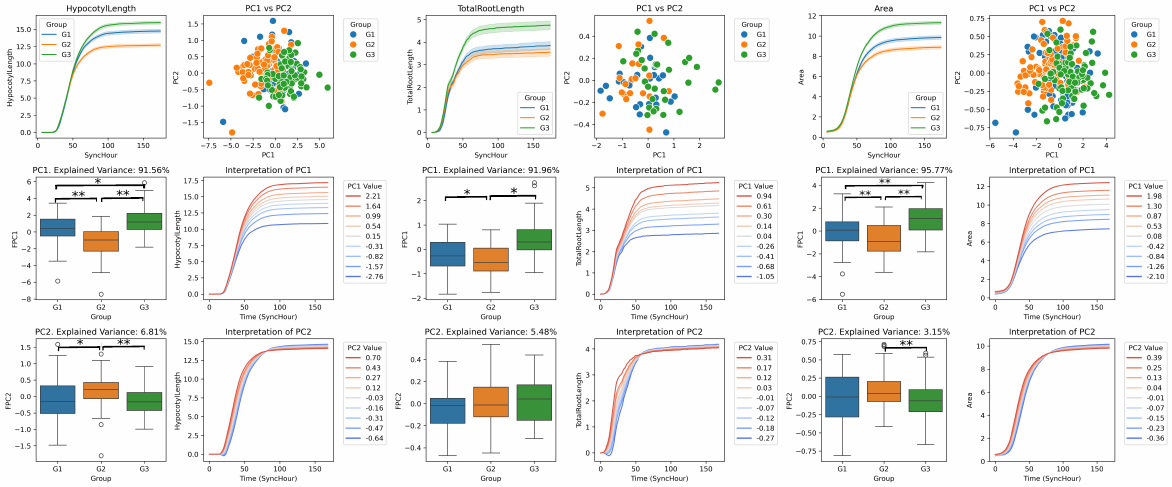}
    \caption{\textbf{Use Case 3 - FPCA Analysis.}To complement the growth dynamics presented in Figure \ref{fig:use_case_3}-B, we performed FPCA to summarize and compare the developmental trajectories of hypocotyl length, total root length, and projected area across genotypes G1, G2, and G3. The first two principal components accounted for over 90\% of the total variance in each trait, capturing the main temporal patterns of growth. 
    For each metric, the top panels show the mean trajectories with standard error bars for each genotype \textcolor{black}{and a scatter plot of PC1 vs PC2}. The middle and bottom rows illustrate the distribution of FPCA scores by group for PC1 and PC2, along with visual interpretations of each component.
    \textcolor{black}{p-values: * < 0.05. ** < 0.001}}
    \label{fig:use_case_3_sup}
\end{figure*}

This appendix provides an intuitive explanation of functional principal component analysis (FPCA) for readers without a quantitative background. The goal is to illustrate, through a simple simulated example, how FPCA decomposes variation across a population of curves into independent modes of variation. We consider curves composed of two distinct components: a smooth, broad parabolic shape and a rapid oscillatory pattern (see Figure \ref{fig:fpca_explanation}. These components vary independently across samples. 

Using a large set of such simulated curves, we apply FPCA to extract the dominant patterns of variation. Each original curve can then be approximately reconstructed as a combination of a mean curve and weighted contributions from the first few principal components. This decomposition helps clarify how variation is structured across a population and which types of patterns dominate.

Fig. \ref{fig:fpca_explanation} shows five example decompositions. Each row corresponds to one simulated curve, split into its two main functional components. This illustration is meant to serve as a visual reference for understanding the role of FPCA in analyzing biological signals that vary smoothly over a continuous domain.

To make this concept more concrete, consider a plant biology scenario where we monitor the growth of plant roots over time. For each plant, we record the length of its primary root at regular intervals, generating a smooth growth curve. These curves reflect dynamic biological processes, including genetic and environmental influences on growth.

Now suppose we are studying several different genotypes or treatments. Each plant’s root grows at its own pace and may exhibit unique features: some may grow rapidly early and then plateau, while others grow steadily or even display fluctuating growth due to stress or environmental factors.

By applying FPCA to this dataset of root growth curves, we can:

\begin{enumerate}
    \item Summarize the dominant patterns of variation: For instance, the first principal component (PC1) might capture differences in overall growth speed (e.g., fast vs. slow growers), while the second component (PC2) might reflect differences in the timing of growth acceleration (e.g., early vs. late spurts).

    \item Reduce dimensionality: Rather than analyzing hundreds of time points, each curve can be represented compactly by just a few scores (weights) corresponding to its projection onto the first few functional components.
    
    \item Cluster or classify plants based on growth patterns: FPCA scores can be used to group plants with similar dynamic traits or to distinguish between genotypes or treatments based on how their roots grow over time.
\end{enumerate}

This approach is particularly valuable in the plant phenotyping scenarios covered by Chronoroot, where growth dynamics are critical but can be challenging to summarize with static metrics. FPCA allows us to capture and quantify subtle temporal trends in a principled, interpretable way, even when the curves are complex or noisy.

\begin{table*}[t!]
\centering
\caption{\textcolor{black}{Detailed report of segmentation performance across plant organs for \textit{arabidopsis thaliana}}.}
\begin{tabular}{lllllllll}
\toprule
Configuration & Metric & Complete Root & Main Root & Lateral Root & Seed & Hypocotyl & Leaf & Petiole \\
\midrule
\multirow{4}{*}{Residual} & DC & 0.800±0.123 & 0.803±0.122 & 0.735±0.176 & 0.686±0.277 & 0.699±0.230 & 0.847±0.162 & 0.758±0.161 \\
 & HD & 6.658±9.703 & 6.649±9.895 & 8.948±11.217 & 14.811±16.148 & 10.323±13.220 & 5.169±7.701 & 7.160±12.339 \\
 & Cp & 0.930 & 0.933 & 0.895 & - & - & - & - \\
 & Cr & 0.937 & 0.939 & 0.897 & - & - & - & - \\
\hline
\multirow{4}{*}{Residual (Fast)} & DC & 0.797±0.126 & 0.800±0.126 & 0.726±0.183 & 0.699±0.256 & 0.695±0.228 & 0.844±0.163 & 0.751±0.162 \\
 & HD & 7.162±10.081 & 7.243±10.218 & 9.931±10.178 & 14.850±15.882 & 10.051±13.968 & 5.348±9.297 & 7.592±13.285 \\
 & Cp & 0.926 & 0.929 & 0.901 & - & - & - & - \\
 & Cr & 0.935 & 0.937 & 0.892 & - & - & - & - \\
\hline
\multirow{4}{*}{Standard} & DC & 0.798±0.119 & 0.800±0.119 & 0.720±0.195 & 0.696±0.281 & 0.695±0.240 & 0.844±0.165 & 0.758±0.156 \\
 & HD & 7.765±12.436 & 7.867±12.513 & 7.750±9.282 & 15.792±17.682 & 8.730±10.412 & 5.147±7.480 & 6.666±12.053 \\
 & Cp & 0.934 & 0.938 & 0.890 & - & - & - & - \\
 & Cr & 0.937 & 0.938 & 0.915 & - & - & - & - \\
\hline
\multirow{4}{*}{Standard (Fast)} & DC & 0.793±0.123 & 0.796±0.123 & 0.711±0.203 & 0.698±0.276 & 0.687±0.245 & 0.840±0.166 & 0.753±0.155 \\
 & HD & 7.902±12.523 & 7.986±12.661 & 8.791±9.679 & 16.349±18.227 & 8.916±11.339 & 7.688±12.846 & 6.953±11.259 \\
 & Cp & 0.929 & 0.933 & 0.874 & - & - & - & - \\
 & Cr & 0.936 & 0.936 & 0.901 & - & - & - & - \\
\bottomrule
\end{tabular}
\label{tab:arabidopsis-detailed-metrics}
\end{table*}
\begin{table*}[ht]
\caption{\textcolor{black}{Detailed report of segmentation performance across plant organs for tomato.}}
\label{tab:tomato-detailed}
\begin{tabular}{lllcccccc}
\toprule
Training & Configuration & Metric & Complete Root & Main Root & Lateral Root & Seed & Hypocotyl & Aerial \\
\midrule
\multirow{4}{*}{Tomato} & \multirow{4}{*}{Standard} & DC & 0.836±0.178 & 0.840±0.178 & 0.643±0.294 & 0.866±0.202 & 0.844±0.197 & 0.778±0.225 \\
 & & HD & 26.470±19.212 & 26.726±19.700 & 27.811±19.615 & 17.744±23.242 & 7.578±13.767 & 12.602±16.797 \\
 & & Cp & 0.920±0.133 & - & - & - & - & - \\
 & & Cr & 0.779±0.230 & - & - & - & - & - \\
\hline
\multirow{4}{*}{Tomato} & \multirow{4}{*}{Standard (Fast)} & DC & 0.805±0.194 & 0.812±0.188 & 0.615±0.289 & 0.884±0.163 & 0.834±0.201 & 0.768±0.235 \\
 & & HD & 32.563±19.591 & 32.483±20.232 & 33.433±19.507 & 17.432±22.671 & 12.784±21.839 & 19.241±24.454 \\
 & & Cp & 0.910±0.157 & - & - & - & - & - \\
 & & Cr & 0.733±0.253 & - & - & - & - & - \\
\hline
\multirow{4}{*}{Tomato} & \multirow{4}{*}{Residual} & DC & 0.882±0.154 & 0.884±0.151 & 0.651±0.279 & 0.857±0.191 & 0.894±0.138 & 0.807±0.222 \\
 & & HD & 19.676±20.152 & 15.238±16.119 & 23.138±20.741 & 18.005±21.034 & 6.021±11.986 & 11.156±17.283 \\
 & & Cp & 0.908±0.153 & - & - & - & - & - \\
 & & Cr & 0.868±0.189 & - & - & - & - & - \\
\hline
\multirow{4}{*}{Tomato} & \multirow{4}{*}{Residual (Fast)} & DC & 0.871±0.165 & 0.876±0.159 & 0.632±0.277 & 0.839±0.202 & 0.885±0.147 & 0.785±0.216 \\
 & & HD & 23.438±20.654 & 18.147±18.024 & 24.354±20.547 & 18.172±20.642 & 6.290±11.842 & 13.965±18.417 \\
 & & Cp & 0.904±0.159 & - & - & - & - & - \\
 & & Cr & 0.853±0.197 & - & - & - & - & - \\
\hline
\multirow{4}{*}{Both} & \multirow{4}{*}{Standard} & DC & 0.853±0.173 & 0.862±0.162 & 0.639±0.324 & 0.885±0.159 & 0.850±0.186 & 0.791±0.218 \\
 & & HD & 25.934±20.458 & 27.324±20.089 & 23.581±18.204 & 15.667±21.435 & 10.579±17.906 & 15.651±15.909 \\
 & & Cp & 0.914±0.141 & - & - & - & - & - \\
 & & Cr & 0.816±0.223 & - & - & - & - & - \\
\hline
\multirow{4}{*}{Both} & \multirow{4}{*}{Standard (Fast)} & DC & 0.842±0.181 & 0.849±0.172 & 0.638±0.312 & 0.887±0.150 & 0.851±0.187 & 0.771±0.234 \\
 & & HD & 25.137±20.132 & 26.765±19.903 & 26.755±18.153 & 14.647±20.058 & 8.937±14.857 & 17.830±17.904 \\
 & & Cp & 0.908±0.156 & - & - & - & - & - \\
 & & Cr & 0.791±0.232 & - & - & - & - & - \\
\bottomrule
\multirow{4}{*}{Both} & \multirow{4}{*}{Residual} & DC & 0.893±0.162 & 0.889±0.157 & 0.673±0.301 & 0.886±0.166 & 0.904±0.144 & 0.854±0.229 \\
 & & HD & 9.943±13.860 & 11.355±14.100 & 15.416±16.647 & 11.604±17.573 & 5.976±12.757 & 14.552±23.488 \\
 & & Cp & 0.916±0.138 & - & - & - & - & - \\
 & & Cr & 0.896±0.177 & - & - & - & - & - \\
\hline
\multirow{4}{*}{Both} & \multirow{4}{*}{Residual (Fast)} & DC & 0.889±0.168 & 0.883±0.169 & 0.664±0.288 & 0.881±0.153 & 0.895±0.143 & 0.859±0.196 \\
 & & HD & 10.218±13.685 & 12.090±13.817 & 17.317±18.033 & 12.134±17.901 & 6.445±13.448 & 13.454±19.100 \\
 & & Cp & 0.905±0.165 & - & - & - & - & - \\
 & & Cr & 0.899±0.172 & - & - & - & - & - \\
\hline
\end{tabular}
\end{table*}

\end{document}